\documentclass[journal]{IEEEtran}

\usepackage{float}

\usepackage{blindtext}
\usepackage{amsmath, amssymb, amsthm} 
\usepackage{stmaryrd}
\usepackage[dvipsnames]{xcolor}
\usepackage{hyperref}

\usepackage{flushend}
\usepackage{color, soul} 
\usepackage{comment}
\usepackage{todonotes}

\usepackage{wrapfig}
\usepackage{siunitx}
\usepackage{bm}

\usepackage[utf8]{inputenc} 
\usepackage[T1]{fontenc}    

\usepackage{hyperref}       
\usepackage{url}            
\usepackage{booktabs}       
\usepackage{amsfonts}       
\usepackage{amssymb,amsmath}
\usepackage{nicefrac}       
\usepackage{microtype}      
\usepackage{makeidx}
\usepackage{graphicx} 
\usepackage[]{algorithm2e}
\usepackage[noadjust]{cite}
 
\DeclareMathOperator*{\argmax}{arg\,max}

\begin{document}
%

\title{
Perceptron Theory Can Predict the Accuracy of Neural Networks
}

\author{Denis Kleyko, Antonello Rosato, E. Paxon Frady, Massimo Panella, and Friedrich T. Sommer\\ 
\thanks{Manuscript received 14 December 2020, revised 17 September  2021, 1 September  2022, and 10 November  2022;  accepted 8 January  2023.
FTS was supported by NIH R01-EB026955.
DK has received funding from the European Union's Horizon 2020 research and innovation programme under the Marie Skłodowska-Curie grant agreement No 839179.
The work of DK was supported in part by the DARPA’s VIP (Super-HD Project) and AIE (HyDDENN Project) programs and by AFOSR FA9550-19-1-0241.
The work of FTS and DK  was also supported in part by Intel's THWAI program.
(\textit{Corresponding author:~Denis Kleyko}.)
}
\thanks{D. Kleyko is with the Redwood Center for Theoretical Neuroscience at the University of California, Berkeley, CA 94720, USA and also with Intelligent Systems Lab at Research Institutes of Sweden, 164 40 Kista, Sweden.
}%
\thanks{A. Rosato and M. Panella are with Department of Information Engineering, Electronics and Telecommunications at the University of Rome ``La Sapienza'', Rome, 00184, Italy.}%
\thanks{E. P. Frady is with the Neuromorphic Computing Laboratory, Intel Labs, Santa Clara, CA 95054 USA.}%
\thanks{F. T. Sommer is with the Redwood Center for Theoretical Neuroscience at the University of California, Berkeley, CA 94720, USA, and also with the Neuromorphic Computing Laboratory, Intel Labs, Santa Clara, CA 95054 USA.}%
\thanks{This article has supplementary material provided by the authors and color versions of one or more figures available at \url{https://doi.org/10.1109/ TNNLS.2023.3237381}.
} 
\thanks{Digital Object Identifier 10.1109/TNNLS.2023.3237381} 
}

\markboth{IEEE TRANSACTIONS ON NEURAL NETWORKS AND LEARNING SYSTEMS}%
{Kleyko \MakeLowercase{\textit{et al.}}: Perceptron Theory}

\maketitle

\begin{abstract}

Multi-layer neural networks set the current state-of-the-art for many technical classification problems. But these networks are still, essentially, black boxes in terms of analyzing them and predicting their performance.
Here, we develop a statistical theory for the one-layer perceptron and show that it can predict performances of a surprisingly large variety of neural networks with different architectures. 
A general theory of classification with perceptrons is developed by generalizing an existing theory for analyzing reservoir computing models and connectionist models for symbolic reasoning known as vector symbolic architectures. Our statistical theory offers three formulae leveraging the signal statistics with increasing detail. The formulae are analytically intractable, but can be evaluated numerically. 
The description level that captures maximum details requires stochastic sampling methods. Depending on the network model, the simpler formulae already yield high prediction accuracy.
The quality of the theory predictions is assessed in three experimental settings, a memorization task for echo state networks from reservoir computing literature, a collection of classification datasets for shallow randomly connected networks, and the ImageNet dataset for deep convolutional neural networks. 
We find that the second description level of the perceptron theory can predict the performance of types of echo state networks, which could not be described previously. 
Further, the theory can predict deep multi-layer neural networks by being applied to their output layer.
While other methods for prediction of neural networks performance commonly require to train an estimator model, the proposed theory requires only the first two moments of the distribution of the postsynaptic sums in the output neurons.
Moreover, the perceptron theory compares favorably to other methods that do not rely on training an estimator model. \\
\phantom{  } \\
\phantom{  } 
\end{abstract}


\section{Introduction}
\label{sect:intro}

Due to its ability to provide data-driven solutions to many previously unsolved problems, machine learning is rapidly changing many areas of our life. At the same time, there is growing demand from society~\cite{HLAIEthics, USAI16} to provide transparency and interpretability of machine learning solutions. For Artificial Neural Networks (ANNs) this requires a deeper understanding of their underlying principles. 
There are many different approaches to this problem, for example, discovering and characterizing structure and dynamics emerging during the training phase, such as geometrical structures of the classifier~\cite{papyan2020prevalence}, a classification behavior called shortcut learning~\cite{GeirhosShortcut2020}, or information content about input against training class~\cite{shwartz2017opening}. 
Another important avenue for understanding ANNs is through studying large ANNs in transfer learning tasks~\cite{BommasaniOpportunities2021}.
This can reveal to what extent input-output mappings are just memorized wholesale, in contrast to the input being dissected for specific parts that should elicit a particular output. 
To provide the transparency of ANN computations, it is crucial to analyze the decisions in trained ANNs and to predict and explain the quality of decisions. One avenue to assess the decision quality is to build models that can predict ANN's performance~\cite{egmont1994quality, UnterthinerPredAcc2020,DeChantPredAcc2019}.\footnote{
In the scope of this study, by performance we refer to ANN's empirical accuracy measured on the test data for the corresponding classification problem.
Interchangeably, the term ``actual accuracy'' is also used when appropriate.
}
The most recent works~\cite{UnterthinerPredAcc2020,DeChantPredAcc2019} train another estimator ANN to perform the prediction.

Here, we propose an alternative approach for predicting the expected accuracy on classification problems for different trained networks including deep networks, as well as echo state networks (ESNs)~\cite{Jaeger12}. 
Our approach does not require training another ANN, rather it is based on the theory of the simple one-layer perceptron, which can describe the last layer of the ANN. 
The perceptron theory generalizes an earlier theory~\cite{Frady17}, proposed for hyperdimensional computing or, synonymously, vector symbolic architectures (HD/VSA) and ESNs with unitary recurrent connections.

The perceptron theory presented here includes formulae for three different levels of prediction accuracy. Building on the original formulation~\cite{Frady17}, we propose two novel formulae that under certain conditions yield predictions with higher accuracy than the original theory.
All formulae leverage the first two moments of the statistics of the postsynaptic sums at the output neurons, but differ in how coarsely the multivariate distribution is approximated. 
The new formula with the highest prediction accuracy captures correlations among dendritic sums of neurons, which makes it hard to evaluate numerically. 
The formulae for lower prediction accuracy neglect such correlations, and can be easily evaluated numerically. The neglecting of correlations introduces a bias to the predictions, which, however, can be removed empirically.  

To assess the quality of the predicted accuracies and to demonstrate their effectiveness in interpreting ANNs, we applied the theory was in three experimental settings.
First, in Section~\ref{sect:esns} we applied it to a type of ESN that was not described by the earlier theory \cite{Frady17}.
Second, in Section~\ref{sect:details:UCI} we used a collection of classification datasets with shallow randomly connected ANNs exploring two strategies of training the perceptron.
Third, in Section~\ref{sect:imagenet} we evaluated the accuracies of $15$ deep convolutional ANNs (CNNs) trained on the ImageNet dataset.
The results show high correlation between the actual accuracies and the predictions. 
Thus, the proposed theory identifies critical features for predicting and comparing performances of networks with different architectures. This not only helps to gain a deeper understanding of the principles at work, but also has practical implications.
For example, for a particular task, the theory can be used to select the best suited network from a set of pretrained networks.

\section{Related work in predicting ANN performance}
\label{sect:related}

Investigations on how to analyze and predict the classification performance of ANNs go back to several decades~\cite{egmont1994quality, feraud2002methodology, baehrens2010explain}. Recent interest in this topic has been in the context of Network Architecture Search (NAS)~\cite{deng2017peephole,istrate2019tapas,elsken2018neural}, i.e., studying the design space of a network and the search strategies in that space~\cite{elsken2018neural}. 
Another domain where prediction performance is crucial, is Knowledge Distillation (KD)~\cite{baker2017accelerating}, and the related studies on how to enhance the performance of simpler models by leveraging more complex ones~\cite{romero2014fitnets, dhurandhar2018improving}.     
The topic is also relevant for explaining and interpreting the overall learning mechanism in multi-layer networks~\cite{MONTAVON20181}, including efforts on visualizations deciphering deep networks~\cite{Zeiler14}.

Here, we particularly build on earlier work on how to predict accuracy, either based on the weights in the ANN~\cite{UnterthinerPredAcc2020, MartinPredicting2021}, or based on the activations of the different layers~\cite{DeChantPredAcc2019}.
In~\cite{UnterthinerPredAcc2020}, the authors trained estimators, which were able to reasonably rank the classification performance of a plethora of CNNs.
In~\cite{DeChantPredAcc2019}, the authors trained a ``meta network'' to predict the correctness of an ANN on an individual input data sample, by using the activations in the hidden layers. 
The results of these studies, as well as~\cite{FixClassifier}, support our approach to bisect ANNs into a transformation and a readout perceptron stages. 
In~\cite{UnterthinerPredAcc2020}, it was observed that parameters of the last fully connected weight layer (i.e., readout perceptron) were among the most informative and frequently used ones.
In~\cite{DeChantPredAcc2019}, the authors reported that for predicting network accuracy, the activations of the last hidden and output layer were the most useful.

While most of the previous work on performance prediction relies on training an estimator model (e.g., an elementary regressor or another ANN), in the context of NAS, there is a demand for maximally reducing computational expenses associated with the search~\cite{ZhouEconas2020} that, in turn, stimulates explorations of simple metrics that do not require training an estimator model.
A recent study~\cite{AbdelfattahZero2021} has empirically investigated several such metrics.
Curiously, while some of the investigated metrics have been proposed in the NAS context~\cite{MellorNeural2021} most of the metrics have appeared first in the context of ANNs' pruning and compression~\cite{LeeSNIP2019, TurnerBlockswap2019,WangWinningTicket2020, tanaka2020pruning}. 
It is worth noting that the principal motivation for developing these metrics is in their practical usage for the above mentioned applications.   
In contrast, here we develop a statistical theory that relies on only a few assumptions with the aim of deepening our understanding of the principles at work in ANNs and without employing any estimator model, which would add yet another black box. 
In contrast, here we develop a statistical theory that relies on only a few assumptions, but does not employ an estimator model, which yet adds another black box. 
Thus, potentially our statistical approach can provide novel insights for explaining the quality of ANNs' decisions such as the possibility to compare networks with different architectures using only the weights of their last layers (see Section~\ref{sect:pred:centroids} for details).

Another interesting approach was introduced recently in~\cite{MartinPredicting2021}. 
It does not rely on training an estimator model. 
Instead, statistical metrics are computed solely from the weights of the ANN and, thus, provide a single score to characterize the trained ANN.  
In contrast, the complete formulation of the perceptron theory needs access to both the ANN's weights and the activation patterns in the last hidden layer to estimate two moments of the statistics of the postsynaptic sums (but see Section~\ref{sect:pred:centroids}). 
We compared the methods in Section~\ref{sect:other:predictions} on predicting networks with different architectures, showing that the perceptron theory provides higher quality than the methods from~\cite{MartinPredicting2021} as well as the most performing metric from~\cite{AbdelfattahZero2021}.

\section{Perceptron theory}
\label{sec:versatile}

The perceptron network is the ancestor of all modern ANNs. A perceptron is a simple artificial neuron introduced by Rosenblatt \cite{Perceptron}, in which inputs are weighted by synapses, added linearly, and converted to a binary output by a threshold transfer function.
Classification problems can be naturally mapped to a network of parallel perceptrons (often referred to as one-layer perceptron) in which each perceptron neuron is responsible for detecting a match between the input and one of the classes. The best match of a given input can be determined by adaptively changing the threshold value in all perceptrons~\cite{rosenblatt1961principles}.
Alternatively, one can replace the original perceptron transfer function by a graded transfer function, such as sigmoid or rectified linear unit, so that the output vector of the network still contains graded match information for the different classes. One can also replace the individual neural transfer functions by a global mechanism for maximum detection among all neurons \cite{steinbuch1963learning}, for example, using winner-take-all competition \cite{amari1977competition,grossberg1988nonlinear,maass2000computational}.
Note that maximum detection was not discussed in the original works of Rosenblatt. However, for solving classification problems with multiple classes such a mechanism is required and  
was introduced early on; see, e.g., \cite{steinbuch1963learning} for a natural generalization of the original perceptron with the winner-take-all mechanism for the case of multiple classes. 
Conversely, modern ANNs use graded neural transfer functions, which enables maximum detection on the output vector of the network. Thus, the analysis of classification with a perceptron-like neural network layer has to take maximum detection into account, it should be based on signal detection theory~\cite{PetersonEtAl1954}.   

\subsection{Perceptron theory in the literature}
\label{sect:ptitl}
Curiously, but understandably given the lack of universality of the one-layer perceptron \cite{minkypappert}, one cannot find the theory of perceptron classification in textbooks. However, this theory has been developed piece by piece in the context of understanding more complicated networks, such as the formation of separable sensory representations \cite{babadi2014sparseness}, symbolic reasoning in HD/VSA\footnote{
For readers interested but not familiar with HD/VSA, we would like to recommend the tutorial-like article~\cite{Kanerva09}, which is probably one of the best starting points facilitating the entrance to the area. 
There is also a detailed treatment of the basics of the area that can be found in~\cite{PlateBook} and more recently in a  two-part comprehensive survey~\cite{KleykoSurveyVSA2021Part1,KleykoSurveyVSA2021Part2}. 
When it comes to specific aspects of the area,~\cite{KleykoComputingParadigm2021} can be consulted for representation of data structures while for similarity-preserving representations and applications to classification problems~\cite{Scalarencoding,frady2021computing,FradyFunctionsNICE2022} and~\cite{HDGestureIEEE,KleykoTradeoffs2018,GeClassificationReview2020} can be looked into, respectively.  
For specific aspects such as representation of data structures, similarity-preserving representations,  and applications to classification problems we recommend consulting~\cite{KleykoComputingParadigm2021};~\cite{Scalarencoding,frady2021computing,FradyFunctionsNICE2022}; and~\cite{HDGestureIEEE,KleykoTradeoffs2018}, respectively.  
}~\cite{PlateTr} and some types of ESNs\footnote{Please refer to Appendix~\ref{sect:capacity:theory}, which introduces a minimalistic  variant of ESN as well as provides additional references. 
}~\cite{Frady17}. These studies demonstrated that complicated neural computations, involving recurrent circuitry and nonlinear stages, can be successfully dissected into two stages: a transformation of input data samples to a new $N$-dimensional space, and a one-layer perceptron that reads out the transformations of input data to classes or network outputs.

In the one-layer perceptron, the synaptic weights are a linear transformation of the inputs to the postsynaptic sums of the neurons. The classification is correct if the postsynaptic sum is the largest in the neuron that corresponds to the actual class of the input data sample.
Up to date, the most complete version of a Gaussian theory of the perceptron was presented in \cite{Frady17}. By generalizing signal detection theory \cite{PetersonEtAl1954}
and building on earlier work on how Gaussian distributions can be transformed \cite{Owen1980}, the work in~\cite{Frady17} shows that if the input data sample belongs to one of $D$ classes, then the predicted accuracy (denoted by $a$), i.e., the probability that the perceptron output ($\mbox{class}_{\texttt{out}}$) is the correct class ($\mbox{class}_{\texttt{inp}}$), is given by
\noindent
\begin{equation}
\begin{split}
a: &= p(\mbox{class}_{\texttt{out}} = \mbox{class}_{\texttt{inp}}) =\\ &\int_{-\infty}^{\infty} \frac{dx}{\sqrt{2\pi}}  \; e^{-\frac{1}{2}{x}^2} \left[ \Phi\left(\frac{\sigma_{r}}{\sigma_{h}}x+\frac{\mu_{h}-\mu_{r}}{\sigma_{h}}\right)\right]^{D-1}.
\end{split}
 \label{eq:pcorr:orig1}
 \end{equation}
 \noindent
Here, $\Phi(x)$ is the cumulative Gaussian; $\mu_{h}$ and $\sigma_{h}$ denote mean and standard deviation of the postsynaptic sum of the output neuron
that corresponds to the correct class; 
$\mu_{r}$ and $\sigma_{r}$ denote the mean and standard deviation of the postsynaptic sum 
for all other neurons. The formula describes the performance of all flavors of HD/VSA models~\cite{PlateTr, HDNP17, Kanerva09,  Gallant13, KleykoHolographic2017, Rachkovskij2001, FradySDR2020} and also describes some variants of ESNs~\cite{ESN03, ESNtut12}. 

\subsection{Novel contribution to Perceptron theory}
\label{sect:octpt}

Intuitively, Eq.~(\ref{eq:pcorr:orig1}) computes the probability that for any given value $x$ of the postsynaptic sum in the output neuron that represents the correct class, the postsynaptic sums in all other neurons are smaller than $x$.
Note that Eq.~(\ref{eq:pcorr:orig1}) makes three strong assumptions about the distributions of the postsynaptic sums:
\noindent
\begin{itemize}
    \item[i)] The distributions are normal.
    \item[ii)] The distributions for ``distractor'' classes are all the same. 
    \item[iii)] The distributions for different classes are independent.
\end{itemize}
\noindent
For analyzing HD/VSA models, these assumptions are justified because information is represented by normalized pseudo-random vectors and the symbolic operations in HD/VSA conserve pseudo-randomness and norm. 
Furthermore, these models do not have weights that are learned from data.
When predicting the accuracy with weights learned from input distributions, for example, in ESNs or ANNs, some of these assumptions will be violated. 

We generalize Eq.~(\ref{eq:pcorr:orig1}) in two steps and demonstrate that the generalization can predict a broader variety of ANN architectures.
Our first step of generalization of Eq.~(\ref{eq:pcorr:orig1}) is to drop assumption ii) and compute individual predicted accuracies for each class $i \in \{1,2,...,D\}$:
\noindent
\begin{equation}
\begin{split}
\textbf{a}_i : &= p(\mbox{class}_{\texttt{out}} = i|\mbox{class}_{\texttt{inp}} = i) = \\
&\int_{-\infty}^{\infty} \frac{dx}{\sqrt{2 \pi} \bm{\sigma}_{i}} e^{-\frac{ (x-\bm{\mu}_{i})^2 }{2 \bm{\sigma}_{i}^2}} \prod_{j=1, j \neq i}^{D-1} \Phi(x,\bm{\mu}_{j}, \bm{\sigma}_{j}) 
\end{split}
 \label{eq:pcorr}
 \end{equation}
\noindent
Here
$\bm{\mu}$ and $\bm{\sigma}$ denote vectors representing the first two moments (mean and standard deviation) of the statistics of the postsynaptic sums for all classes.
The assumptions for Eq.~(\ref{eq:pcorr}) to hold are reduced to i) and iii).

Our second generalization is to drop assumption iii), that is, we allow the postsynaptic sums for different neurons to be correlated according to a multivariate normal distribution:\footnote{
We assume that the correct class is always in the first position of $\textbf{x}$ (i.e., $\textbf{x}_1$).
This is done just for convenience to make the writing of the next integrals in Eq.~(\ref{eq:pcorr:mvn}) more straightforward.
}   
\noindent
\begin{equation}
\begin{split}
&\textbf{a}_i:=p(\mbox{class}_{\texttt{out}} = i|\mbox{class}_{\texttt{inp}} = i)= \\
&\int_{-\infty}^{\infty} d\textbf{x}_1
\int_{-\infty}^{\textbf{x}_1} d\textbf{x}_2
\dots 
\int_{-\infty}^{\textbf{x}_1} d\textbf{x}_D
\; {\cal N}(\textbf{x}, \bm{\mu}, \bm{\Sigma}),
\end{split}
 \label{eq:pcorr:mvn}
 \end{equation}
\noindent
where ${\cal N}(\textbf{x}, \bm{\mu}, \bm{\Sigma})$ is the multivariate normal distribution with the full covariance matrix $\bm{\Sigma}$ replacing the variance vector $\bm{\sigma}$ for independent Gaussians in Eq.~(\ref{eq:pcorr}), see Appendix~\ref{sect:capacity:theory} for connection between Eq.~(\ref{eq:pcorr:mvn}) and Eq.~(\ref{eq:pcorr}). 
The assumptions for Eq.~(\ref{eq:pcorr:mvn}) to hold are reduced to i).

To aggregate the individual class accuracies in $\textbf{a}$ into the overall prediction, we form the expectation over all $D$ classes:
\noindent
\begin{equation}
\sum_{i=1}^{D} \textbf{f}_i \textbf{a}_i,
\label{eq:aggr:acc}
 \end{equation}
\noindent
where $\textbf{f}_i$ is the prior probability of $i$th class in the data. 
Recall that vector $\textbf{a}$ stores predicted accuracies for all $D$ individual classes where the predictions are obtained according to one of the above formulations of the perceptron theory (cf. Eqs.~(\ref{eq:pcorr:orig1})-(\ref{eq:pcorr:mvn})). 
This single prediction score characterizes the accuracy of the network for the whole classification task.
The prior probability of each class $\textbf{f}_i$ is estimated using the frequency of $i$th class labels in the empirical samples of the data such as the training or test ones. 
The predicted accuracy can be compared to the actual accuracy of the network.

\section{Experiments}
\label{sect:perf}

\subsection{Predicting the performance of echo state networks}
\label{sect:esns}

First, we test our generalized perceptron theory on predicting an ESN performance in the trajectory association task~\cite{PlateBook}. In Fig.~\ref{fig:ESN:acc}, we compare two ESNs with different readout perceptrons.\footnote{
Please see Appendix~\ref{sect:capacity:theory} for the detailed description of the experiment and the types of ESNs being used. 
}
The synapses of one perceptron (blue line) are set to the centroids of inputs of the different classes (codebook-based).
The synapses in the other perceptron (red line) have been optimized with linear regression (regression-based). The average predicted accuracy for the both versions is plotted in Fig.~\ref{fig:ESN:acc} as a function of the delay of the input occurred in the sequence of input vectors.

\begin{figure}[tb]
\centering
\includegraphics[width=0.9\linewidth]{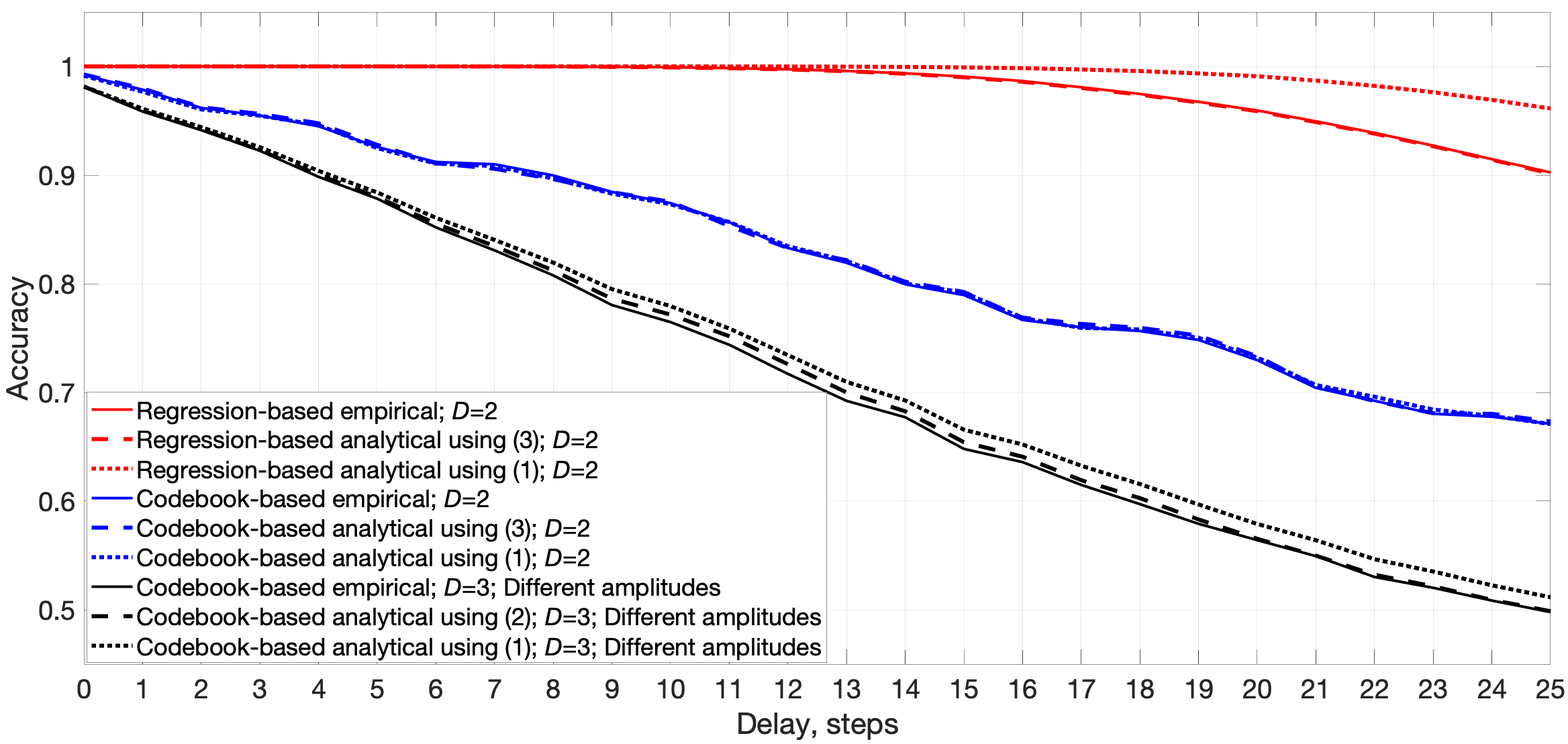}
\caption{
The accuracy of the ESN against delay for the case of codebook-based and regression-based readout perceptrons.
The following values for the ESN parameters were used: $N=100$, $D=2$ (for red and blues lines) or $D=3$ (for black lines), $\kappa=4$.
The length of test sequences was $10,000$. 
All reported values were averaged over $50$ simulations with pseudo-random codebooks. 
}
\label{fig:ESN:acc}
\end{figure}
The solid lines depict empirical accuracies. The dotted lines are the predictions by theory in Eqs.~(\ref{eq:pcorr:orig1})-(\ref{eq:pcorr:mvn}), using the statistics reported in Fig.~\ref{fig:ESN:stat} in the Appendix. 
The original formulation in Eq.~(\ref{eq:pcorr:orig1}) correctly describes the experimental performance of the perceptron with centroid filters but overestimates the performance of the perceptron with regression filters (i.e., it has some bias). 
The bias occurs because Eq.~(\ref{eq:pcorr:orig1}) neglects the correlation between the linear filters of the regression-based readout, visible in the right panel of Fig.~\ref{fig:ESN:stat}.
The dashed lines in Fig.~\ref{fig:ESN:acc} depict the predictions by Eq.~(\ref{eq:pcorr:mvn}), the accuracies of both networks are predicted correctly.

The predictions by Eq.~(\ref{eq:pcorr}) are omitted because they did not differ from the predictions of Eq.~(\ref{eq:pcorr:orig1}) when $D=2$. To see differences between these two formulations, we run another experiment with $D=3$ and a perceptron with centroid filters where the inputs had different amplitudes (black solid line).  
For this experiment, Eq.~(\ref{eq:pcorr:orig1}) (black dotted line) overestimated the accuracy, while Eq.~(\ref{eq:pcorr}) (black dashed line) provided accurate predictions\footnote{We also studied the perceptron theory predictions for a synthetic binary classification problem in the case of non-independent distributions of postsynaptic sums in Appendix~\ref{sect:pred:corr}. 
}.

\subsection{Predicting the performance  of shallow networks}
\label{sect:details:UCI}

We also applied the theory in Eq.~(\ref{eq:pcorr}) to shallow feedforward randomly connected ANNs~\cite{RVFLorig}, which are similar to ESNs without the recurrent connections. 
The evaluation was done on a collection of the classification tasks where the data are provided in the form of the extracted features. 
In particular, the reported results are based on $121$ real-world classification datasets obtained from the UCI Machine Learning Repository~\cite{Dua:2019}.
The considered collection of datasets has been initially analyzed in a large-scale comparison study of different classifiers and the interested readers are kindly referred to the original work~\cite{HundredsClassifiers} for more details.
The only preprocessing step was to normalize features in the range $[0, 1]$.

For the sake of brevity, here we do not go into the details of the transformation stage. 
The interested readers are kindly referred to~\cite{intRVFL}.
In essence, the transformation resembles Random Vector Functional Link (RVFL) networks~\cite{RVFLorig}, which is a particular variation of shallow feedforward randomly connected ANNs.
Moreover, it is very similar to the known approaches of using HD/VSA for classification~\cite{KussulDiagnostics1998, goltsev2005combination, Rasanen2015tr, BICA16, KleykoIndustrial2018, HDGestureIEEE, DiaoGLVQHD2021, HuangSpeaker2022 }.

The search of the hyperparameters for each dataset has been done according to~\cite{HundredsClassifiers} using the grid search over $\lambda$ (regularization parameter), $N$,  and $\kappa$ (activation function parameter).
$N$ varied in the range $[50, 1500]$ with step $50$;  $\lambda$ varied in the range $2^{[-10, 5]}$ with step $1$; and $\kappa$ varied in the set of $\{1,3,5,7\}$.
The obtained optimal hyperparameters were used to estimate the cross-validation accuracy on all datasets.

Similar to the previous section, we considered two ways of forming the perceptron. 
The first way is common in HD/VSA.  
It forms a linear filter as a centroid of a class by simply superimposing all transformations of input data for this class.  
The perceptron then is simply the collection of all the centroids.
Formally, let us assume that the transformations for all training data are stored in the matrix $\mathbf{X}$ and the corresponding class labels are stored in the ground truth vector $\mathbf{y}$; then the centroid for $i$th class (denoted as $\mathbf{W}_i$) is obtained as the sum of the corresponding transformations in $\mathbf{X}$:  
\noindent
\begin{equation}
\mathbf{W}_i = \sum_{j, \mathbf{y}_j=i }  \mathbf{X}_j.
 \end{equation}
\noindent
In addition, it is common to normalize each centroid to unit norm: 
\noindent
\begin{equation}
\mathbf{W}_i = \frac{ \mathbf{W}_i } { || \mathbf{W}_i ||_2  }.
 \end{equation}

The second way is common in shallow feedforward randomly connected ANNs
The perceptron is the result of the ridge regression, which uses the transformations of training data and the ground truth class labels to obtain the optimal values of the perceptron $\mathbf{W}$ as:
\begin{equation}
\mathbf{W}= 
\mathbf{Y}^{\top} \mathbf{X} (\mathbf{X}^{\top} \mathbf{X} + \lambda \mathbf{I})^{-1},
\label{eq:readout:regr}
\end{equation}
\noindent 
where $\mathbf{I}$ denotes the identity matrix, $\lambda$ is a regularization parameter used in the ridge regression, and $\mathbf{Y}$ is the one-hot representation of the ground truth from $\mathbf{y}$. 
Thus, the main difference between the considered approaches in forming the perceptrons is that the regression filters are obtained using the optimization procedure while the centroid filters are non-parametric. 

Across the datasets, the average cross-validation accuracy for the perceptron with centroid filters was $0.70$ while that for the perceptron with regression filters was $0.80$.
The Pearson correlation coefficient  between the obtained result was $0.80$.  
It is clear that the perceptrons obtained from the ridge regression usually outperformed the ones with the centroids.

\begin{figure}[tb]
\centering
\includegraphics[width=0.95\columnwidth]{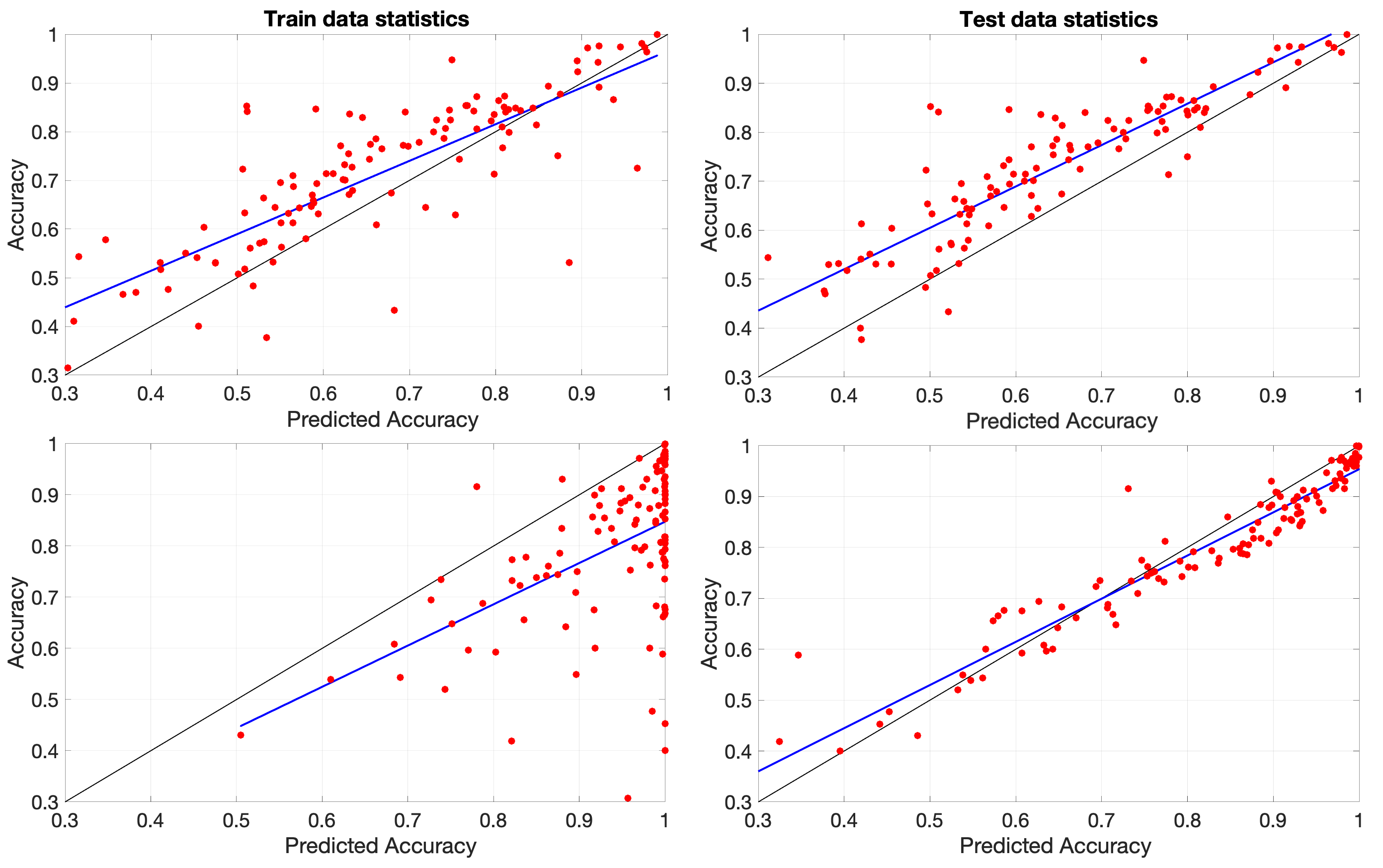}
\caption{
The cross-validation accuracy  against the predicted accuracy where the predicted accuracy for individual classes were calculated according to Eq.~(\ref{eq:pcorr}).  
Each point corresponds to a dataset.
The upper panels: the perceptron with centroid filters.
The lower panels: the perceptron with regression filters. 
}
\label{fig:intrvfl:cent}
\end{figure}

The upper panels in Fig.~\ref{fig:intrvfl:cent} present the cross-validation accuracies against their corresponding predicted accuracies for the perceptron with centroid filters.   
The upper-left panel in the figure corresponds to the case when the statistics for Eq.~(\ref{eq:pcorr}) was obtained from the training data. 
The upper-right panel in the figure corresponds to the case when the statistics for Eq.~(\ref{eq:pcorr}) was obtained from the test data.
The Pearson correlation coefficient between the accuracy and predicted accuracy calculated from the statistics for the training data was $0.79$ while that of the test data was $0.90$. The usage of the statistics for the test data improved the quality of the predictions.
The reduced correlation for the case when the statistics for the training data was used is expected and happens due to the ``leakage'' of the data. 
Since the same data samples are used to both form the perceptron and to estimate the statistics of the postsynaptic sums, the obtained estimates might differ significantly from the ones observed for previously unseen data samples (i.e., test data).

The lower panels in Fig.~\ref{fig:intrvfl:cent} present the cross-validation accuracies against their corresponding predicted accuracies for the perceptron with regression filters. 
The Pearson correlation coefficient between the actual accuracy and predicted accuracy for the statistics from the training data was $0.47$ while that for the test data was $0.97$. 
Similar to the results for the perceptron with centroid filters, the usage of the statistics for the test data improved the quality of the predictions.
The difference, however, was that the predicted accuracies from the training data statistics did not correlate well with the accuracies. 
The most likely explanation is that in the case of the ridge regression, the perceptron is calculated to maximize the accuracy on the training data. 
As a side effect, the statistics of the postsynaptic sums is too promising and so many networks are predicted to achieve high accuracy on the test data.
Other notable differences were that compared to the perceptron with centroid filters the errors and bias are smaller for the ridge regression with test data statistics.
For instance, after removing the bias  the mean value of the absolute error was $0.03$ for the perceptron with regression filters while for the perceptron with centroid filters it was $0.12$.

\begin{figure}[tb]
\centering
\includegraphics[width=0.9\columnwidth]{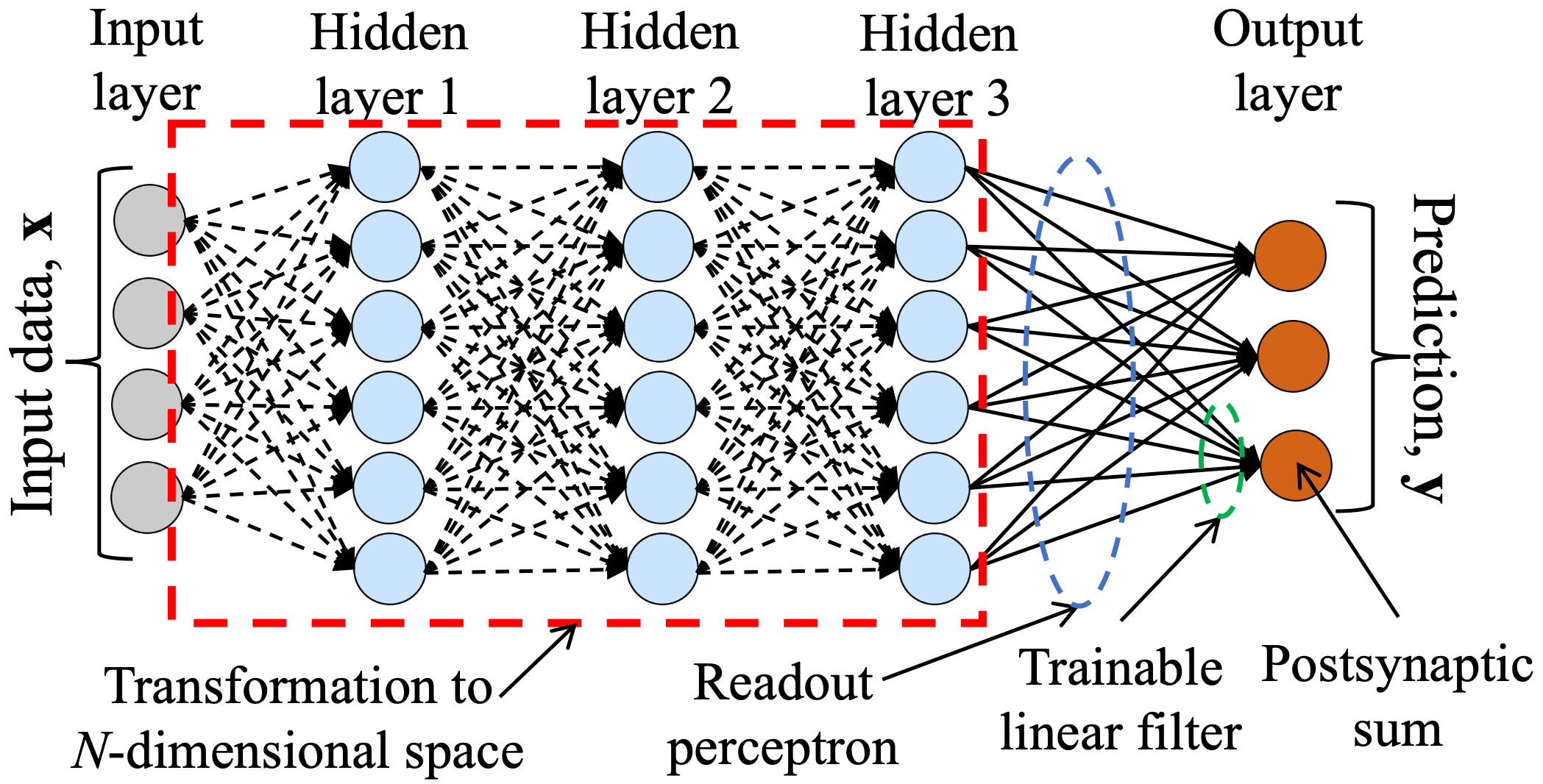}
\caption{
Dissecting the functionality of an ANN into multi-layer transformation and perceptron stages.
The figure indicates components of the network by the terms used in the article. 
}
\label{fig:ann:dissect}
\end{figure}

Thus, the main finding here is the sequent: for both perceptrons with centroid and regression filters, the theory in Eq.~(\ref{eq:pcorr}) introduced some bias but the Pearson correlation coefficients between the predictions and the actual accuracies were high: $0.90$ and $0.97$, respectively. 
These differences should be attributed to the effect of the assumptions used in Eq.~(\ref{eq:pcorr}). 
The next section will elaborate more on this topic.

\subsection{Predicting the performance of deep CNNs}
\label{sect:imagenet}

\begin{figure*}[tb]
\centering
\includegraphics[width=1.6\columnwidth]{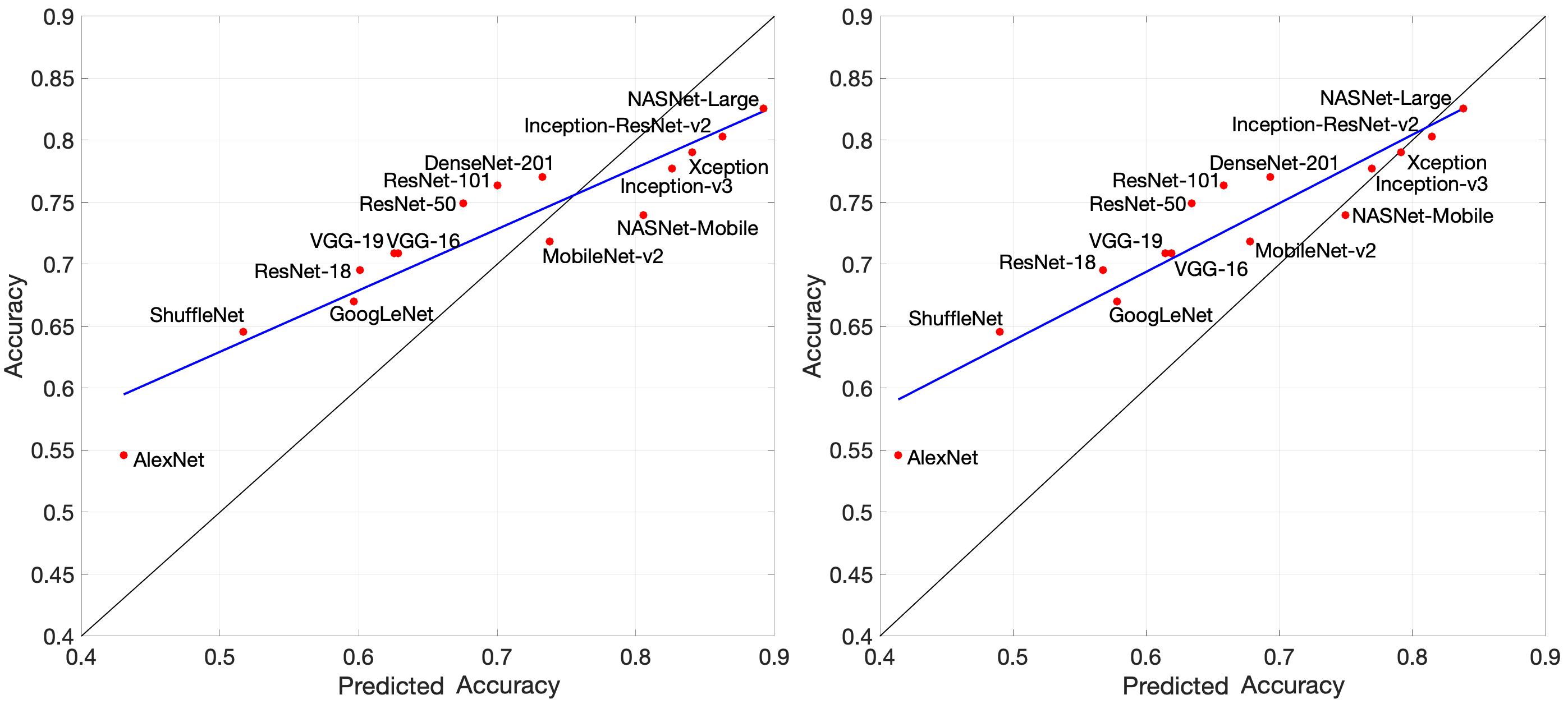}
\caption{
The accuracy of $15$ deep CNNs on the ILSVRC 2012 validation dataset against the predicted accuracy.
Left panel: the individual predicted accuracies were calculated according to Eq.~(\ref{eq:pcorr}).  
Right panel: the individual predicted accuracies were calculated similar to Eq.~(\ref{eq:pcorr}) but using the kernel distributions (Pearson correlation coefficient was $0.94$).
}
\label{fig:in:pcorr:norm}
\end{figure*}

\begin{figure}[tb]
\centering
\includegraphics[width=0.7\columnwidth]{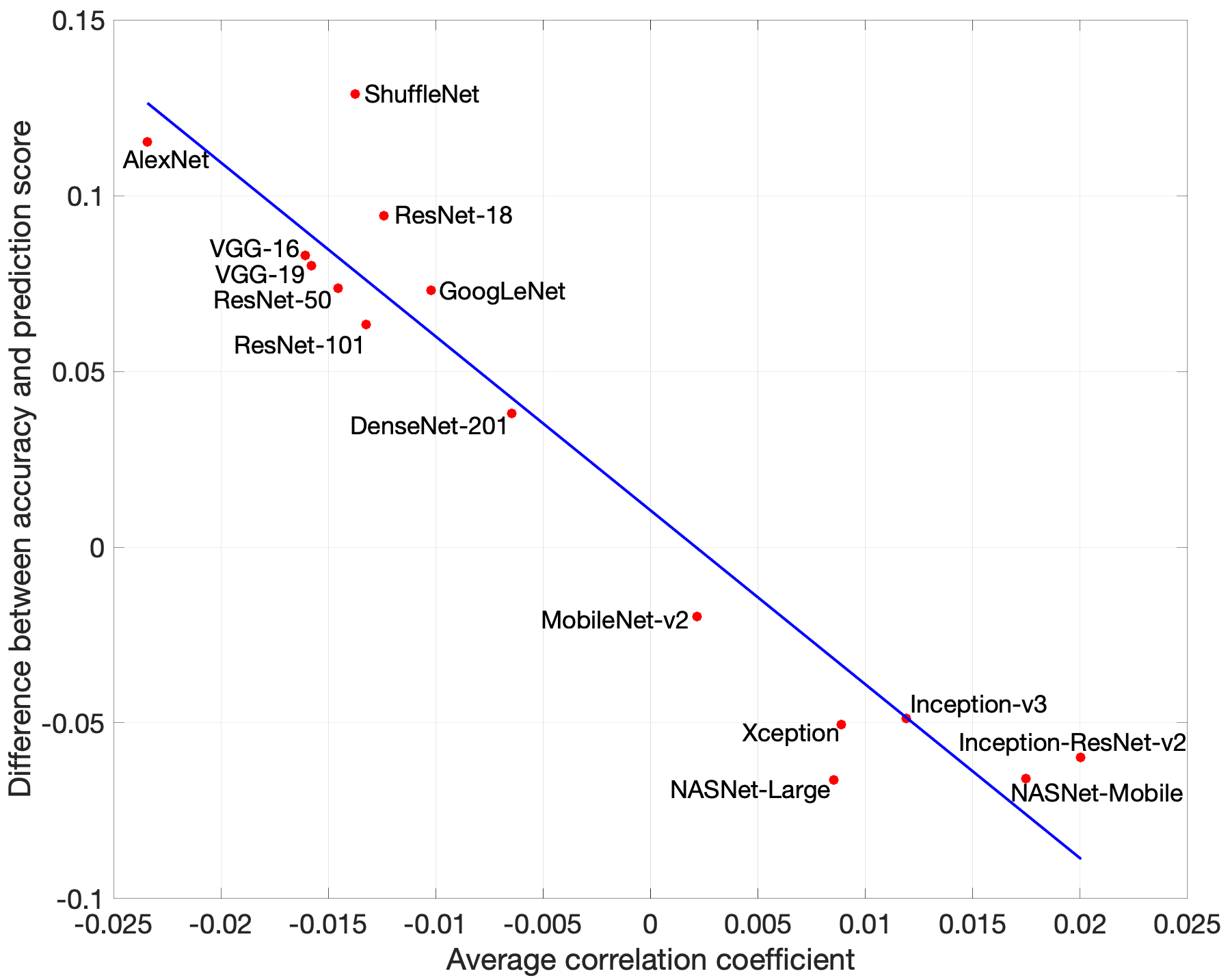}
\caption{
The difference between the accuracy and the predicted accuracy for $15$ deep CNNs against the average Pearson correlation coefficient in the covariance matrices. 
The predicted accuracies were calculated according to Eq.~(\ref{eq:pcorr}). 
}
\label{fig:in:pcorr:covar}
\end{figure}

\subsubsection{Deep ANNs in terms of transformation and perceptron stages}
\label{sect:imagenet:inerpretation}
In the remainder, we will apply the perceptron theory described in Section~\ref{sec:versatile} on deep ANNs. 
To apply the theory, we dissect the holistic functionality of a deep network into two parts: a multi-layer transformation stage and a single-layer classification by a perceptron as depicted in Fig.~\ref{fig:ann:dissect}.
It is not very common to look at ANNs this way but, for example, a recent work~\cite{papyan2020prevalence} has used the same dissection to study the terminal phase of the training.
Note that this dissection is conceptual, it affects neither training nor inference processes.
This bipartite view suggests that most of the network is doing a transformation, i.e., produces a useful representation of an input data sample.
As we will see below, a similar conceptual dissection can be applied if the transformation stage includes convolutional layers (i.e., CNNs) or recurrent connections (i.e., RNNs), as long as the last hidden layer and the output later are densely connected.
Finally, note that using the final outcomes of the multi-layer transformation stage (i.e., activations of the last hidden layer) does not mean that any information about the network is lost because obtaining these activations requires propagating the activity through all the layers that precede the last hidden one (cf. dashed rectangle in Fig.~\ref{fig:ann:dissect}).

\begin{figure*}[tb]
\begin{center}
\begin{minipage}[h]{1.4\columnwidth}
\centering
\center{\includegraphics[width=0.9\columnwidth]{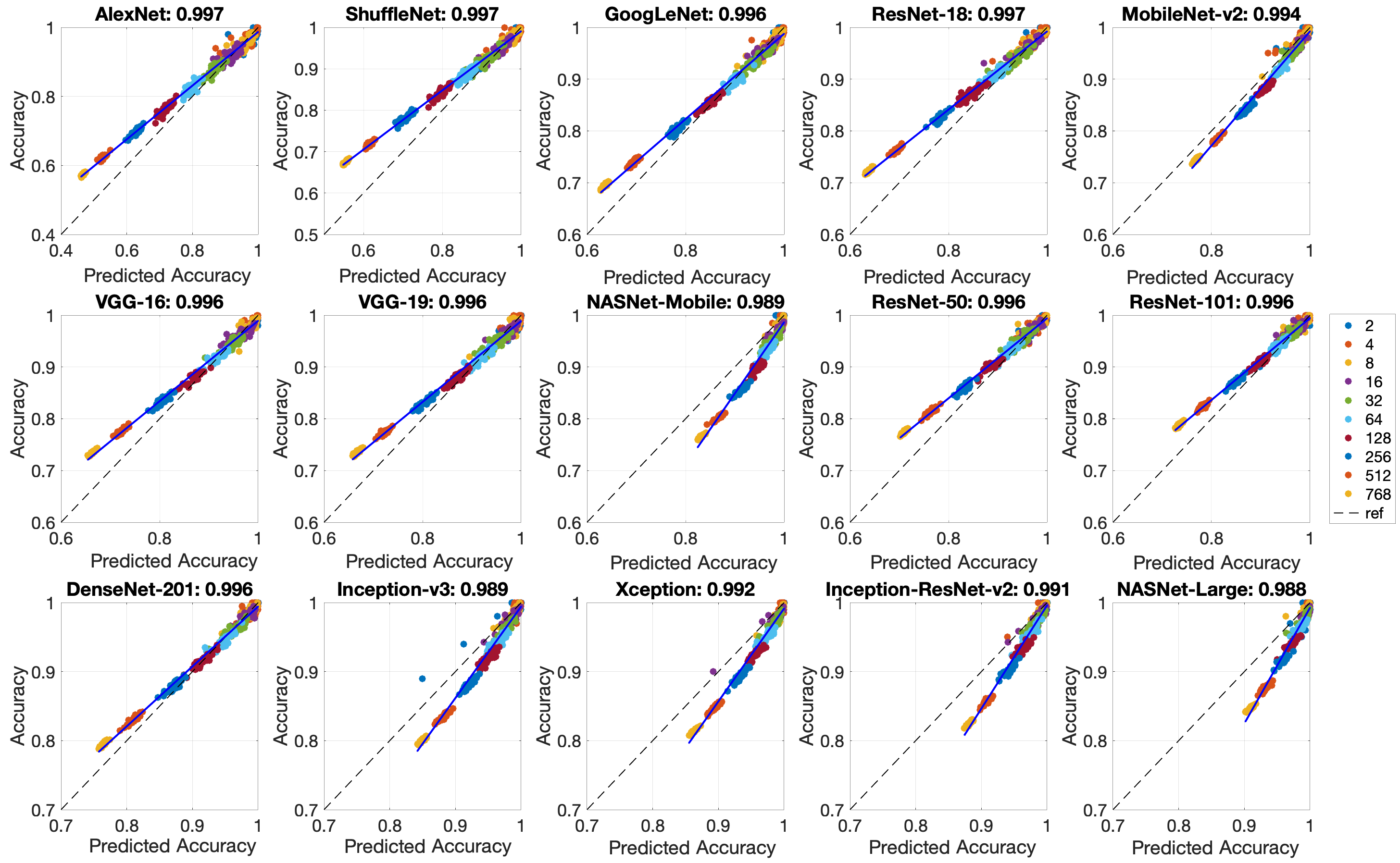}}
\end{minipage}
\hfill
\begin{minipage}[h]{0.58\columnwidth}
\center{\includegraphics[width=1.0\columnwidth]{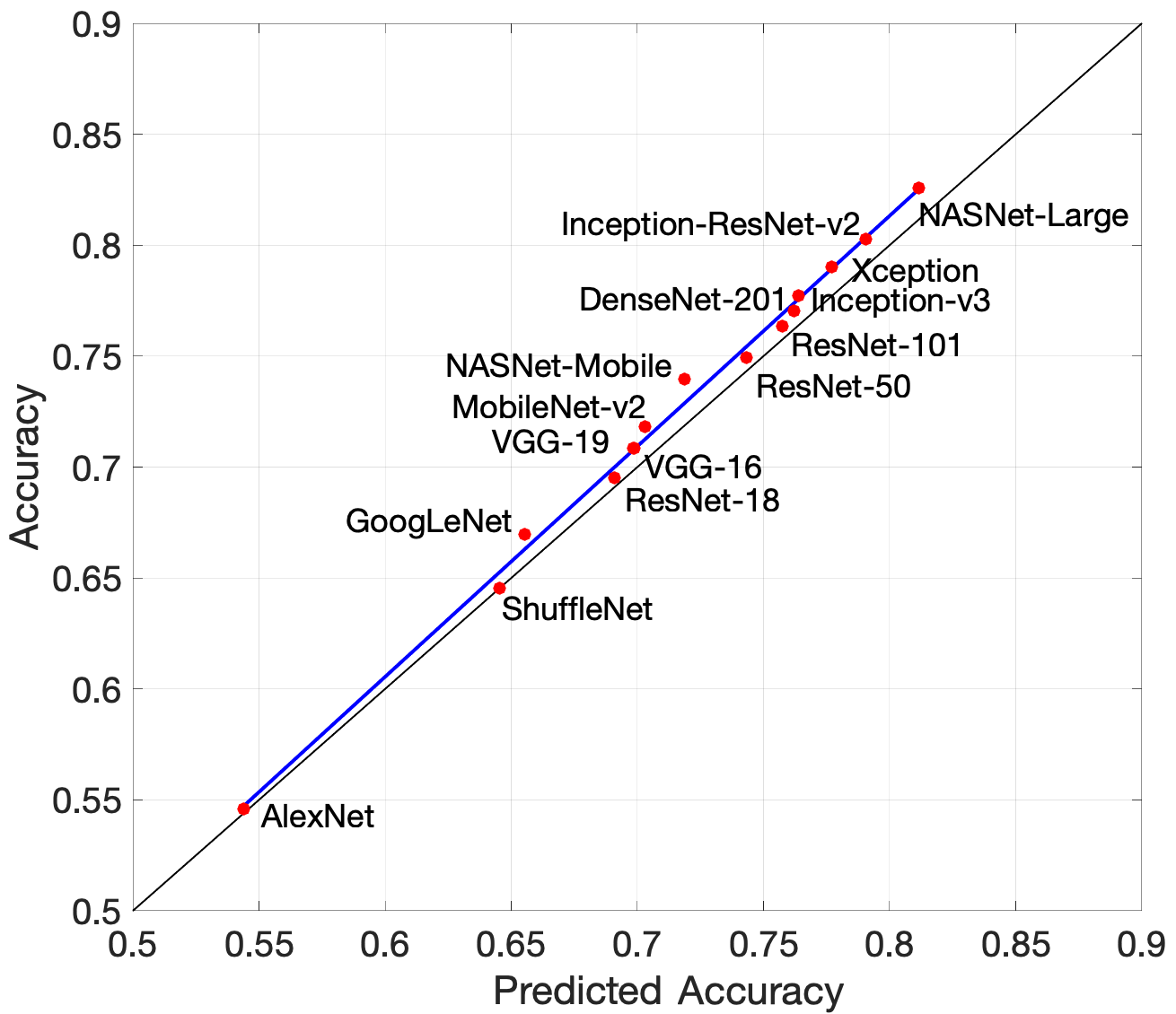}} 
\end{minipage}
\end{center}
\caption{
The accuracy of $15$ deep CNNs on the ILSVRC 2012 validation dataset against the predicted accuracy.
Left: sub-problems of different size randomly sampled from ImageNet.
The predicted accuracies for individual classes were calculated according to Eq.~(\ref{eq:pcorr}).  
Color of point indicates the size of the sub-problems. 
Title for each panel states network's name and the Pearson correlation coefficient.
Right: The predicted accuracies of networks were calculated according to Eq.~(\ref{eq:pcorr}) and then compensated using the lines from the left panel.
}
\label{fig:in:pcorr:norm:sub}
\end{figure*}

\begin{figure}[tb]
\centering
\includegraphics[width=0.8\columnwidth]{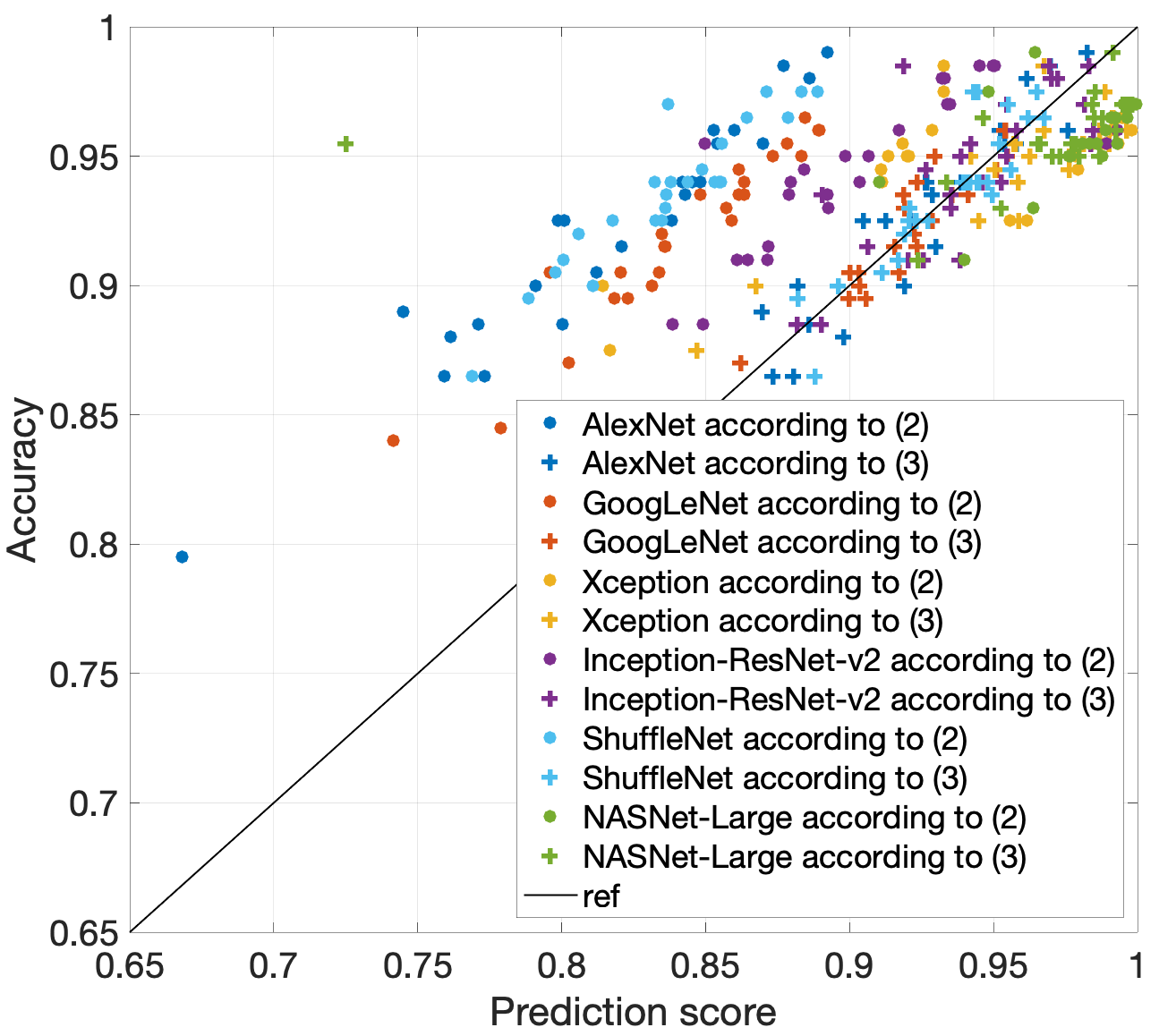}
\caption{
The accuracy of $6$ deep CNNs against the predicted accuracy for the sub-problems of size $4$.
The predicted accuracies for individual classes were calculated either according to Eq.~(\ref{eq:pcorr}) or Eq.~(\ref{eq:pcorr:mvn}). 
}
\label{fig:in:mvn}
\end{figure}

\subsubsection{Setup}
\label{sect:imagenet:setup}

Here, we describe the results of experiments with a set of well-known deep CNNs, which were pretrained on the ImageNet dataset~\cite{ImageNet}.
The ImageNet dataset arose from ImageNet Large Scale Visual Recognition Challenge (ILSVRC). 
Currently, it is one of a well-established benchmarks in the object category classification and detection.
ImageNet includes $D=1,000$ classes corresponding to different object categories, which makes it hard to preform well on this task. 
The training data of ImageNet includes over $14$ million images.
The validation set from ILSVRC 2012 challenge has $50,000$ images in total; exactly $50$ images per each class.

In the experiments below, $15$ pretrained deep CNNs were used:
AlexNet~\cite{AlexNet},
GoogLeNet~\cite{GoogLeNet},
ResNet-18,
ResNet-50,
ResNet-101~\cite{ResNet},
VGG-16,
VGG-19~\cite{VGG},
ShuffleNet~\cite{ShuffleNet}, 
MobileNet-v2~\cite{MobileNet}, 
DenseNet-201~\cite{DenseNet},   
Inception-v3~\cite{Inception},
Inception-ResNet-v2~\cite{InceptionResNet}, 
Xception~\cite{Xception}, 
NASNet-Mobile, and 
NASNet-Large~\cite{NASNet}. 
The actual accuracy of the deep CNNs was assessed using the validation set from ILSVRC 2012 challenge. Blue solid lines in the figures below are the lines fitted to the results.

For each network, we first saved the weight matrix corresponding to the connections between the last hidden layer and the output layer. 
Recall that in our interpretation this weight matrix (i.e., readout perceptron) is treated as a set of $D$ linear filters where for each class the linear filter is the corresponding $N$-dimensional vector.
Second, for all networks we have also obtained the activations of the last hidden layer for each image in the ILSVRC 2012. 
Recall that these activations are seen as the results of the multi-layer transformation stage of the network. 
All linear filters and the last hidden layer activations of a particular class were used to calculate the postsynaptic sums.
Note that in cases when it is easier to obtain the statistics in the form of activations of ANN's last layer, they can be used to directly compute the required statistics. 
The resultant matrix was used as a data sample to estimate moments of probability density functions, which are involved in calculating the predicted accuracy $\textbf{a}_i$ for that class.

\subsubsection{Predictions according to Eq.~(\ref{eq:pcorr})}
\label{sect:imagenet:pcorr}

We compared the actual accuracy of each network against its predicted accuracy, where the predicted accuracies for individual classes were calculated according to Eq.~(\ref{eq:pcorr}) and then aggregated into a single prediction (left panel in Fig.~\ref{fig:in:pcorr:norm}).\footnote{
Practically, it was averaging as in the ILSVRC 2012 each class has the same number of images so $f_i=1/D=10^{-3}$.  
}
The results were mixed. 
On one hand, the Pearson correlation coefficient between the actual accuracies and predicted accuracies was $0.93$,  indicating a high similarity between the actual and predicted accuracies. 
Also Kendall's $\tau$ correlation coefficient measuring the quality of the ranking of the networks performances was $0.86$, which is rather high.
For instance, it is worth noting here that if one would just use the number of flops for each network (which is a meaningful information for deep CNNs) as the performance prediction, these correlation coefficients are only $0.50$ and $0.49$, respectively.  
On the other hand, however, there are obvious issues with the predicted accuracies.
First, there is a clear bias in the predicted accuracies.
Second, single points deviate noticeably from the fitted linear trends.
For example, even after compensating for bias the largest error was observed for AlexNet, where the predicted accuracy overestimated the accuracy by $0.05$; the mean value of the absolute error was $0.02$.

One possible explanation of these prediction errors is that assumption i) in Eq.~(\ref{eq:pcorr}) is violated, the assumption that the postsynaptic sums are distributed normally. We quantified the effect of this assumption by using non-parametric kernel density estimation for the probability density functions and then calculated the predicted accuracy in the same way as in Eq.~(\ref{eq:pcorr}).  
The accuracies predicted with the kernel estimation (right panel in Fig.~\ref{fig:in:pcorr:norm}), indeed, differ somewhat from the ones depicted in the left panel. However, dropping the assumption of Gaussianity did not significantly reduce the error in predicted accuracies.
The strong bias was still present and the deviations of the points from the fitted line were still non-negligible. The mean value of the absolute error after compensating the bias was at $0.02$. 
In Section~\ref{sect:other:predictions:all}, we also provide the predictions produced by other prediction methods.

The next natural step was to investigate the effect of assumption iii), that the distributions of individual postsynaptic sums are all independent.

\subsubsection{On removing bias introduced by Eq.~(\ref{eq:pcorr})}
\label{sect:imagenet:bias}

Since Eq.~(\ref{eq:pcorr}) does not take into account the covariance matrix, it is worth checking whether some information contained in the covariance matrix itself is able to explain the bias introduced by the predicted accuracies. 
Fig.~\ref{fig:in:pcorr:covar} depicts the difference between the observed accuracy and the predicted accuracy against the average value of Pearson correlation coefficients in the covariance matrices of each network.
We see that the difference closely follows the average values of Pearson correlation coefficients for different networks. 
In particular, the Pearson correlation coefficient between the differences and the average Pearson correlation coefficients was $-0.95$.
Thus, at least for the ImageNet dataset we can conclude that it is possible to use the average Pearson correlation coefficient to predict whether the predicted accuracy is too optimistic or too pessimistic about the actual accuracy.



Morevoer, since it is hard to conclude much from only $15$ datapoints, we used ImageNet to create sub-problems with smaller numbers of classes. 
In our experiments, the sub-problem of a given size was generated by randomly choosing (without replacement) the classes to be included in the sub-problem.
The following sizes of sub-problems were used $\{2, 4, 8, 16, 32, 64, 128, 256, 512, 768\}$.
For each network and for each sub-problem size, $40$ different sub-problems were evaluated.

Each panel in Fig.~\ref{fig:in:pcorr:norm:sub} (left) corresponds to a deep CNN. Each point corresponds to a sub-problem with the size of the sub-problem represented by color. 
First of all, we see that each network develops its own bias, which is in line with the bias present in the left panel in Fig.~\ref{fig:in:pcorr:norm}.
This suggests that assuming independence between distributions of postsynaptic sums of output neurons caused a noticeable bias effect on the predicted accuracies.
Moreover, the error between the actual and predicted accuracies caused by the bias was increasing with the size of the sub-problem. 
However, what is astonishing is that the Pearson correlation coefficients between the actual and predicted accuracies were almost exactly one for individual networks (the lowest one was $0.99$ for NASNet-Large).
This is another implicit indication that the assumption of normal distribution is not critical for predicting accuracy.   
Moreover, the lines fitted for each network can now be used to compensate the corresponding predicted accuracies in the left panel in Fig.~\ref{fig:in:pcorr:norm}.
For the compensated predicted accuracies against the actual accuracies the Pearson correlation coefficient was $0.998$ (right panel in Fig.~\ref{fig:in:pcorr:norm:sub}).
We also noticed that the compensations based on the sub-problems on individual networks have almost removed bias and unsystematic deviations between the accuracies. 
The compensated predicted accuracies overestimate the actual accuracies to a lesser degree than that depicted in Fig.~\ref{fig:in:pcorr:norm}, e.g., the largest error (overestimated) was $0.02$ for NASNet-Mobile.

 \begin{figure}[tb]
\centering
\includegraphics[width=0.80\columnwidth]{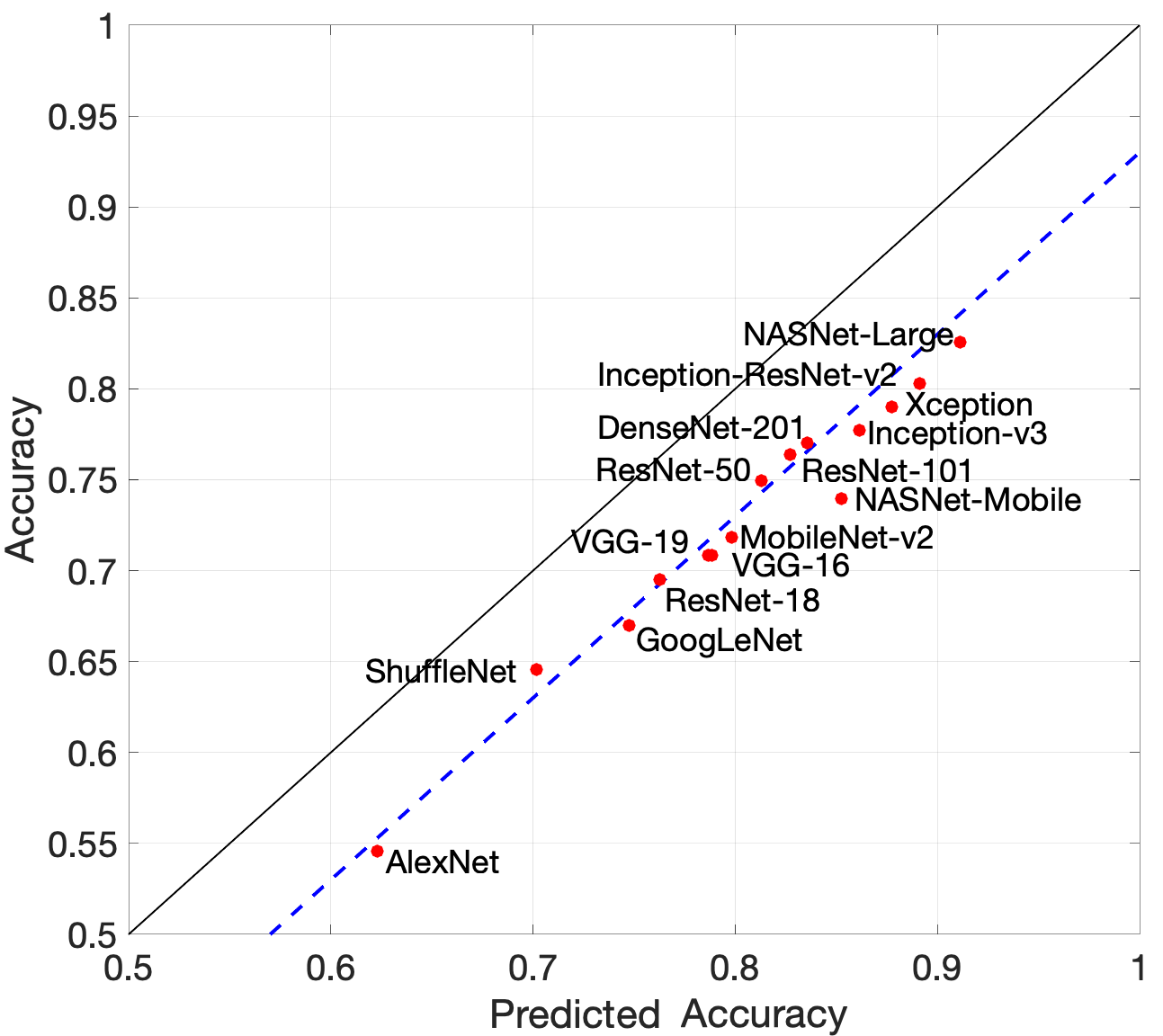}
\caption{
The accuracy of $15$ deep CNNs against the predicted accuracy.
The predicted accuracies were calculated according to Eq.~(\ref{eq:pcorr:mvn}) using Monte Carlo sampling.
}
\label{fig:in:pcorr:MC}
\end{figure}

\begin{figure}[tb]
\centering
\includegraphics[width=0.9\columnwidth]{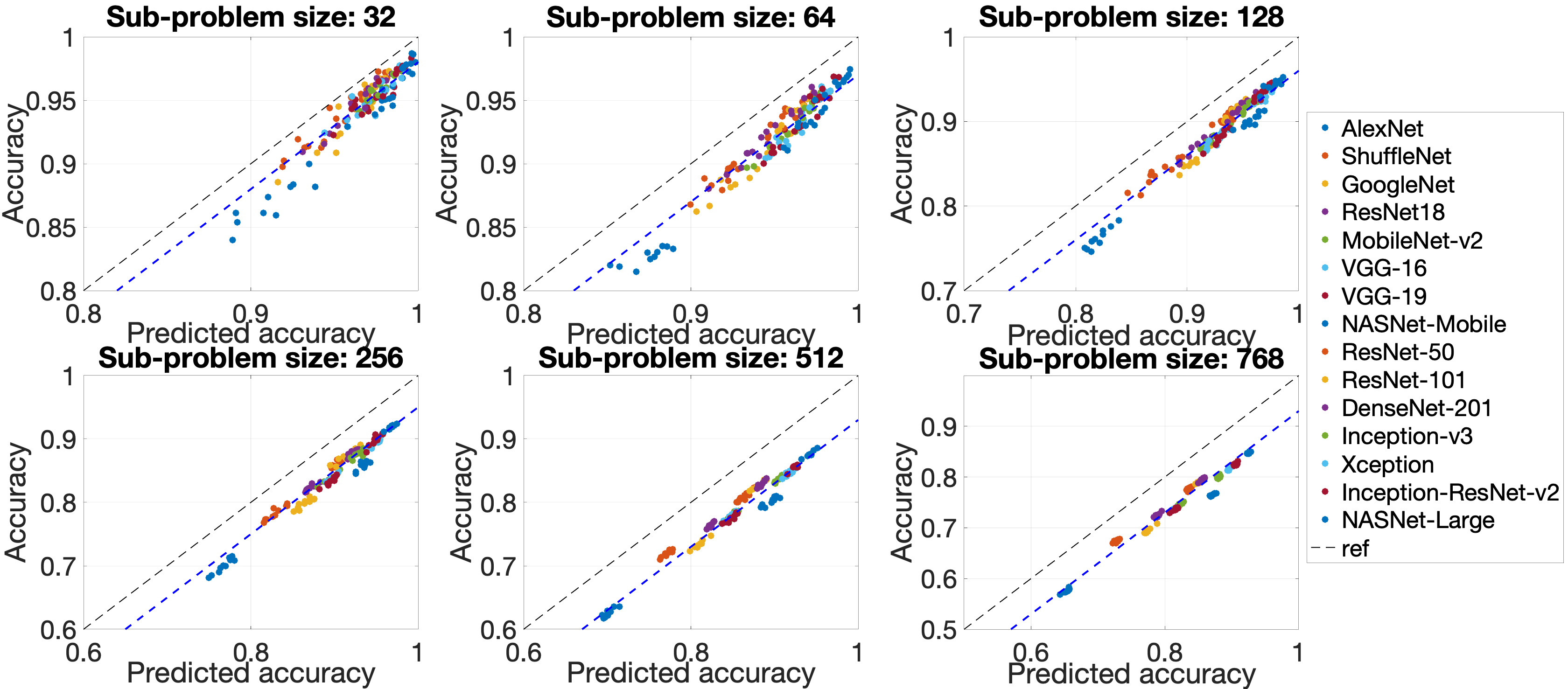}
\caption{
The accuracy of $15$ deep CNNs against the predicted accuracy. Each panel corresponds to a fixed size of a sub-problem.
The predicted accuracies were calculated according to Eq.~(\ref{eq:pcorr:mvn}) using Monte Carlo sampling.
}
\label{fig:in:pcorr:MC:sub}
\end{figure}

\subsubsection{Predictions according to Eq.~(\ref{eq:pcorr:mvn})}
\label{sect:imagenet:pred:full}

The results in the right panel of Fig.~\ref{fig:in:pcorr:norm:sub} are encouraging from the point that the compensated predicted accuracies nearly perfectly corresponded to the accuracies. 
Obviously, Kendall's $\tau$ correlation coefficient was $1.00$.
On the other hand, in order to make the compensation it is necessary to observe accuracies of smaller sub-problems, which is highly undesirable from the practical applicability point of view.
The solution to this problem is to get rid of the independence assumption by calculating the predicted accuracies according to Eq.~(\ref{eq:pcorr:mvn}).
This approach, however, has its own complications as the numerical integration of Eq.~(\ref{eq:pcorr:mvn}) is challenging, even for the moderate number of classes.
Thus, to demonstrate that Eq.~(\ref{eq:pcorr:mvn}) addresses the bias issue, we have performed another experiment. We selected the sub-problems from ImageNet where the size of the sub-problem was fixed to $4$ classes. 
Different from the previous experiment, we first randomly chose $10,000$ sub-problems and calculated both the actual and predicted accuracies using Eq.~(\ref{eq:pcorr}). 
Then we handpicked $25$ most challenging sub-problems choosing the ones whose predicted accuracies calculated using Eq.~(\ref{eq:pcorr}) resulted in large errors. For these sub-problems, we used both formulae, Eq.~(\ref{eq:pcorr}) and Eq.~(\ref{eq:pcorr:mvn}), to predict the accuracies (see Fig.~\ref{fig:in:mvn}). 
For the considered networks there was no bias in the predictions calculated according to Eq.~(\ref{eq:pcorr:mvn}), which is an empirical demonstration that Eq.~(\ref{eq:pcorr:mvn}) can predict a wider range of network architectures.

\begin{figure}[tb]
\centering
\includegraphics[width=0.8\columnwidth]{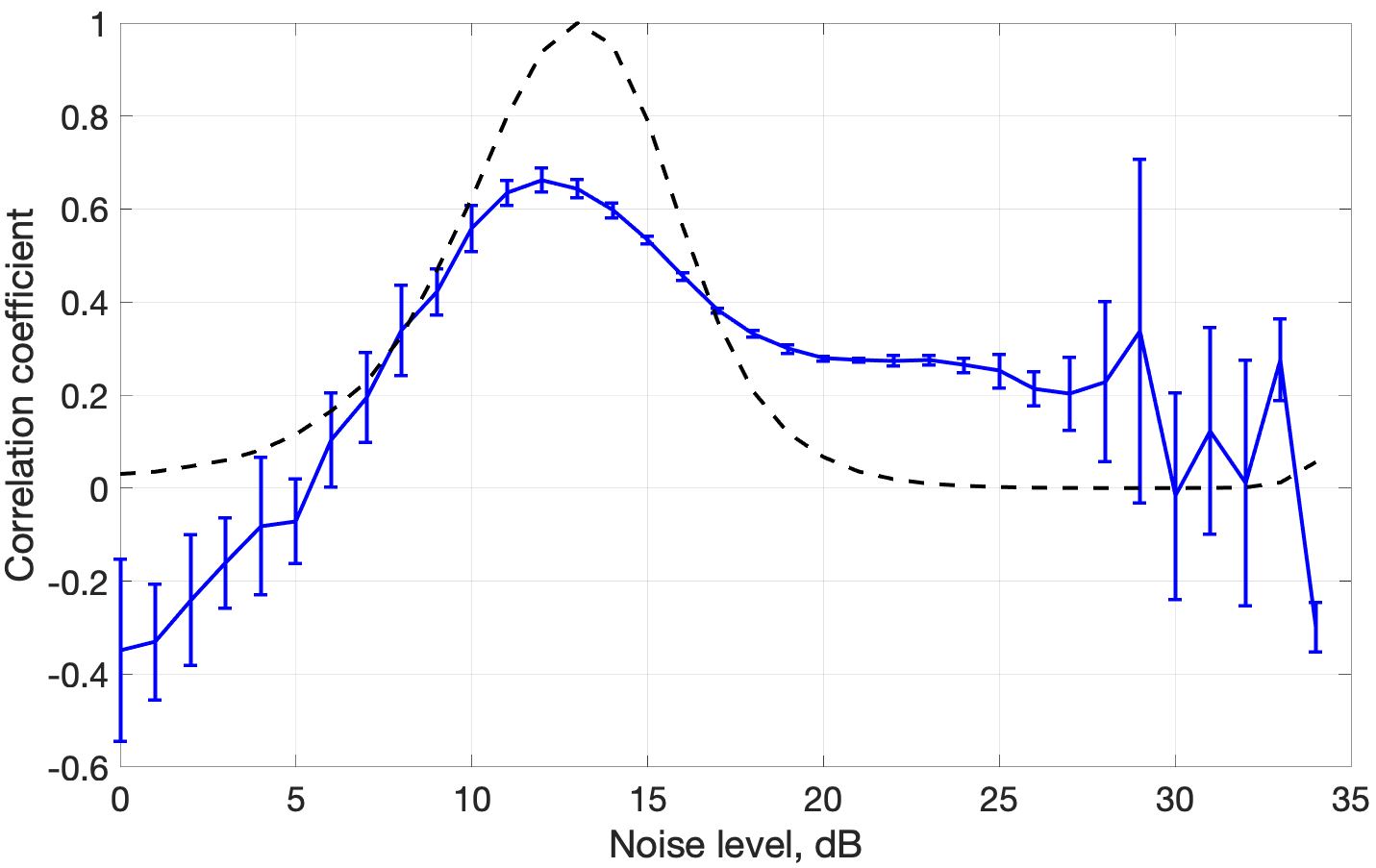}
\caption{
The solid line depicts the mean Pearson correlation coefficients for $15$ deep CNNs between their accuracy and the predicted accuracies calculated using only the perceptron when disturbing the linear filters with white noise for different level of noise. The bars depict standard deviations. 
The dashed line depicts standard deviations of the predicted accuracies normalized by the largest value.  
}
\label{fig:in:pcorr:noise}
\end{figure}

As Fig.~\ref{fig:in:mvn} suggests, calculating the predicted accuracies with Eq.~(\ref{eq:pcorr:mvn}) should address the problem of bias introduced by Eq.~(\ref{eq:pcorr}). However, the numerical integration for problems with many classes such as ImageNet is not tractable. 
We partially circumvented this problem by using a Monte Carlo approach for sampling from the estimated distributions.
The results obtained with the Monte Carlo approach (see Fig.~\ref{fig:in:pcorr:MC}) were quite obviously better than the ones obtained with Eq.~(\ref{eq:pcorr}).
The Pearson correlation coefficient was $0.98$, the Kendall's $\tau$ correlation coefficient was $0.92$. 
Another nice outcome was that the points were arranged on a line (dashed line) parallel to the line of perfect correspondence (solid line), i.e., all the predictions slightly overestimated the actual accuracies by a constant offset. 
This offset just comes from the fact that one should use the normalization constant in our estimations, which Monte Carlo sampling does not provide.

Fig.~\ref{fig:in:pcorr:MC:sub} presents the results of 
applying the same sampling approach to different sizes of sub-problems randomly formed from ImageNet. 
Overall, we still see the same trend as in Fig.~\ref{fig:in:pcorr:MC}, i.e., that the predicted accuracies were approximately offset from the line of perfect correspondence by a constant (dashed lines). 
The difference is more noticeable with the decreased size of a sub-problem, which indicates the increased importance of normalization constants.

\begin{figure*}[tb]
\centering
\includegraphics[width=1.8\columnwidth]{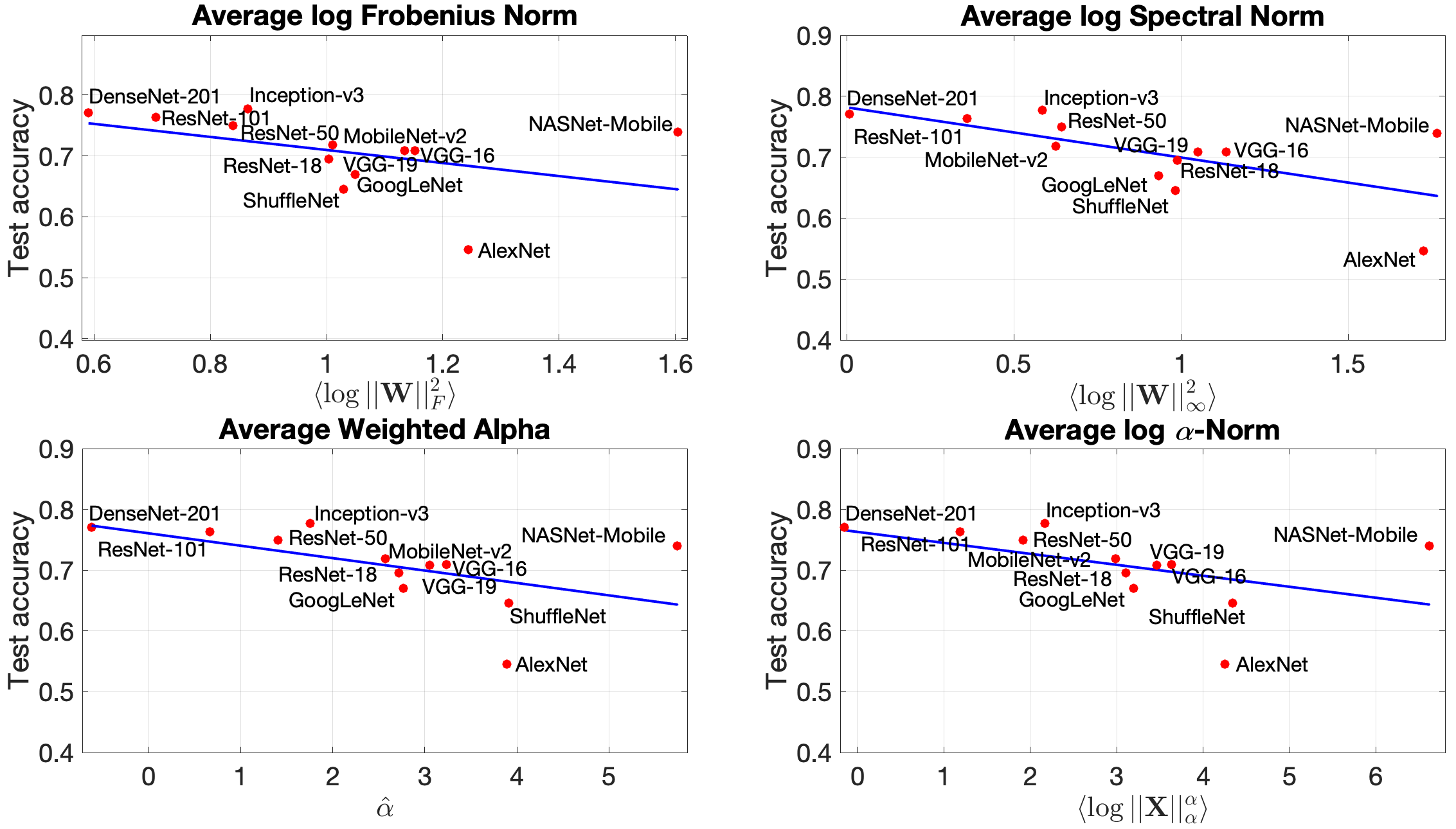}
\caption{
The accuracy of $12$ deep CNNs on the ILSVRC 2012 validation dataset against the prediction metrics from~\cite{MartinPredicting2021}.
The prediction metrics were obtained using full sets of networks' weights.
}
\label{fig:weight:watcher:all}
\end{figure*}

\subsubsection{Predicted accuracy from the readout perceptron only}
\label{sect:pred:centroids}

One interesting question to the proposed theory is: how well could it work given only the readout perceptron but no activations of the last hidden layer?
Quite surprisingly, as we will see below, it was possible to achieve a $\tau$ correlation coefficient of $0.71$ by computing the predicted accuracy only from the readout perceptrons, without any access to the transformations of input data samples.

Intuitively, a possible way to calculate the required statistics (i.e.,  $\bm{\mu}$ and $\bm{\sigma}$) would be to use linear filters themselves in place of the hidden layer activations.
The issue with this approach, however, is that the postsynaptic sums to linear filters themselves are much higher than that of the hidden layer activations from real data since in reality the activations of the hidden layer will never match perfectly their corresponding linear filters, which means that realistic postsynaptic sums are lower. 
Thus, this situation presents the case when we obtain the maximally possible postsynaptic sum, which is not expected in reality where hidden layer activations will not be the exact copies of the corresponding linear filter. 
Therefore, the predicted accuracies calculated in this way are nearly $1$. 
A possible way to mitigate this issue is to disturb the linear filters by adding some white noise to them. 
The noisy versions of a particular linear filter are used in place of activations of the last hidden layer by the samples of the corresponding class. 
The postsynaptic sums are calculated using the original linear filters and their noisy versions.
The obtained postsynaptic sums are used to estimate the required statistics for $\bm{\mu}$ and $\bm{\sigma}$. 
Thus, the white noise added to the linear filters acts as a way to obtain surrogates of the actual statistical distribution of the data. 

We have made the experiments for different levels of noise. 
In a single experiment, each linear filter was disturbed $50$ times. 
The disturbed versions were used as the hidden layer activations for linear filter's class. 
Fig.~\ref{fig:in:pcorr:noise} (solid line) reports the mean Pearson correlation coefficients between the actual accuracy and the predicted accuracies obtained from surrogate statistics from ten experiments for each level on noise.
The results suggest that either high or low noise levels did not result in high correlation but there was a window between $8$ and $17$ dB where the correlation peaked getting as high as $0.66$.

\begin{figure*}[tb]
\centering
\includegraphics[width=1.8\columnwidth]{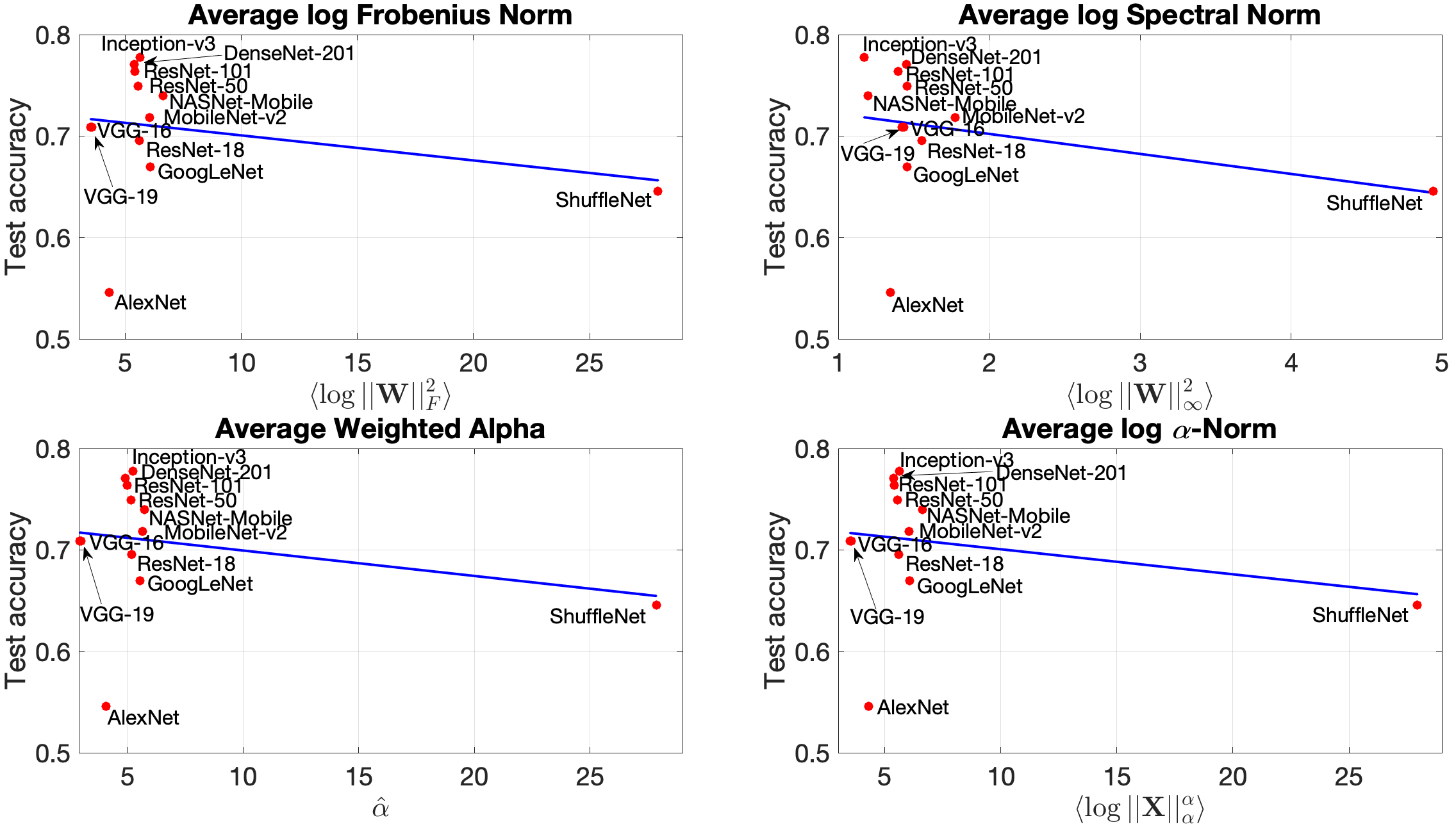}
\caption{
The accuracy of $12$ deep CNNs on the ILSVRC 2012 validation dataset against the prediction metrics from~\cite{MartinPredicting2021}.
The prediction metrics were obtained using only the weights of the readout perceptrons.
}
\label{fig:weight:watcher:last}
\end{figure*}

A natural question to ask would be: how to know which level of noise to use to obtain the highest correlation? 
We think that it could be identified in a straightforward manner; we use two observations to do so. 
First, we expect that different networks have somewhat different accuracy. 
Second, we know that the best value of noise is somewhere in between the extremes. That is because when the noise is too high the predicted accuracies of all the networks would vanish to a random guess while when the noise is too low the predicted accuracies would saturate at one, as explained above. 
Using these observations, we expect that the best level of noise should result in the largest dispersion of the predicted accuracies, which can be simply measured by the standard deviation of the predicted accuracies. 
The dashed line in Fig.~\ref{fig:in:pcorr:noise} depicts mean (across ten experiments) standard deviations for each level of noise. 
As we can see from comparing this curve with the curve for the Pearson correlation coefficients, both curves peaked approximately in the same window between $8$ and $17$ dB.
Thus, when choosing the level of noise corresponding to the largest standard deviation we expect to get close to the peak in the Pearson correlation coefficient. 
The absence of a clear peak might be interpreted as the indication that the accuracy of all the networks is so close to each other that we cannot discriminate between them using only information about the perceptron.  

\begin{table}[t!]

\caption{Quality of predictions obtained from the complete networks.} 
\label{tab:complete:anns}
\centering
\begin{tabular}{|c|c|c|c|}
\hline
Method & Source & Pearson's $r$ & Kendall's $\tau$    \\
\hline
Average log Frobenius norm & \cite{MartinPredicting2021} & $0.43$ & $0.42$ \\ 
\hline
Average log spectral norm & \cite{MartinPredicting2021} & $0.65$ & $0.42$ \\
\hline
Average weighted alpha & \cite{MartinPredicting2021} & $0.52$ & $0.52$ \\
\hline
Average log $\alpha$-norm & \cite{MartinPredicting2021} & $0.48$ & $0.52$ \\
\hline

log \texttt{synflow} & \cite{AbdelfattahZero2021} & $0.66$ & $0.71$ \\
\hline
\hline
Perceptron theory, Eq.~(\ref{eq:pcorr}) & our & $0.90$ & $0.79$\\
\hline
Perceptron theory, Eq.~(\ref{eq:pcorr:mvn}) & our & $\mathbf{0.98}$ & $\mathbf{0.88}$\\
\hline
\end{tabular}

\end{table}

Finally, we can use the obtained predicted accuracies in order to rank networks using Kendall's $\tau$ correlation coefficient to measure the quality of the ranking.
The largest mean $\tau$ was $0.71$, which is not a perfect ranking ($\tau=1$), but it is far better than random ranking ($\tau=0$).
Perhaps not surprisingly, these results are not as good as the results obtained for real  hidden layer activations but they still suggest that the perceptron itself contains rich information about possible performance of a network, which is in line with the observations in~\cite{UnterthinerPredAcc2020}. 
For predictions obtained from the readout perceptron by other estimation methods please refer to Section~\ref{sect:other:predictions:readout}.

\subsubsection{Existing prediction methods for ANNs}
\label{sect:other:predictions}

This section provides the sense on how the predictions produced by the perceptron theory compare to predictions by other existing methods.
As indicated in Section~\ref{sect:related}, most of the related methods rely on training estimator models.
While these methods are useful in applications such as NAS, they hardly provide novel insights into the principles underlying ANNs. 
Notably, however, there are recent studies~\cite{MartinPredicting2021,AbdelfattahZero2021} that do not require any training of the estimator model, similar to our perceptron theory case. 
Therefore, we compared our results to several metrics proposed in~\cite{MartinPredicting2021} and to the best performing metric in~\cite{AbdelfattahZero2021}. All the metrics proposed in~\cite{MartinPredicting2021} are based on the weights of the trained ANN.
While the investigation in~\cite{AbdelfattahZero2021} was concerned with searching for simple ``zero-cost'' metrics that would release the computational burden of performing NAS where the metrics were inspired from the literature related to ANNs' weights pruning~\cite{LeeSNIP2019, tanaka2020pruning,WangWinningTicket2020}. 
For the sake of brevity, we omit the details of calculating the considered metrics, but refer to the corresponding equations in~\cite{MartinPredicting2021} and~\cite{AbdelfattahZero2021}:

\begin{table}[tb]
\caption{Quality of predictions obtained from the readout perceptrons.}
\label{tab:perceptron:anns}
\centering
\begin{tabular}{|c|c|c|c|}
\hline
Method & Source & Pearson's $r$ & Kendall's $\tau$    \\
\hline
Average log Frobenius norm & \cite{MartinPredicting2021} & $0.25$ & $0.09$; \\ 
\hline
Average log spectral norm & \cite{MartinPredicting2021} & $0.31$ & $0.30$ \\
\hline
Average weighted alpha & \cite{MartinPredicting2021} & $0.26$ & $0.06$ \\
\hline
Average log $\alpha$-norm & \cite{MartinPredicting2021} & $0.25$ & $0.09$ \\
\hline
\hline
Perceptron theory, Eq.~(\ref{eq:pcorr}) & our & $\mathbf{0.66}$ & $\mathbf{0.61}$\\
\hline
\end{tabular}
\end{table}

\begin{itemize}
    \item  average log Frobenius norm (denoted as $\langle \log ||\mathbf{W}||_F^2 \rangle $; see Eq.~(3) in~\cite{MartinPredicting2021}),
    \item  average log spectral norm (denoted as $ \langle \log ||\mathbf{W}||_{\infty}^2 \rangle $; see Eq.~(4) in~\cite{MartinPredicting2021}),
    \item  average weighted alpha (denoted as $\hat{\alpha}$; see Eq.~(10) in ~\cite{MartinPredicting2021}),
    \item average log $\alpha$-norm (denoted as $ \langle \log ||\mathbf{X}||_{\alpha}^{\alpha} \rangle  $; see Eq.~(11) in~\cite{MartinPredicting2021}), and
    \item \texttt{synflow}; see Eq.~(1) in~\cite{AbdelfattahZero2021}).   
\end{itemize}
\noindent
Among the metrics considered in~\cite{AbdelfattahZero2021}, we use \texttt{synflow} since it was empirically demonstrated to perform most consistently across all the alternatives. 
A small peculiarity was that we found that using $\log(\texttt{synflow})$ significantly improved Pearson's $r$ so it was used to report the results. 
Since not all of the networks considered in this study were available in the software distribution for reproducing~\cite{MartinPredicting2021}, in the experiments below we limited the ANNs to the subset of twelve deep CNNs (Xception, NASNet-Mobile, and NASNet-Large were excluded).

\paragraph{Predictions obtained from complete networks}
\label{sect:other:predictions:all}

First, to provide the comparison to the perceptron theory predictions reported in Fig.~\ref{fig:in:pcorr:norm} and~\ref{fig:in:pcorr:MC}, we obtained the metrics using full sets of weights for each ANN.
Fig.~\ref{fig:weight:watcher:all} presents the results where each panel corresponds to a metric. 
Table~\ref{tab:complete:anns}\footnote{
Strictly speaking, the Pearson correlation coefficients for all metrics from~\cite{MartinPredicting2021} are negative.
However, for the sake of presentation clarity, their absolute values are reported.
The same note applies to the results reported in Table~\ref{tab:perceptron:anns}.  
} 
summarizes the quality of predictions in terms of two correlation coefficients used above that were calculated with the actual accuracies: Pearson's $r$ and Kendall's $\tau$. 
Besides the statistical metrics from~\cite{MartinPredicting2021} and \texttt{synflow}, the table also reports the coefficients for the corresponding predictions (i.e., excluding three CNNs mentioned above) obtained for the perceptron theory according to Eq.~(\ref{eq:pcorr}) (see Fig.~\ref{fig:in:pcorr:norm}) and Eq.~(\ref{eq:pcorr:mvn}) (see Fig.~\ref{fig:in:pcorr:MC}).

Amongst the statistical metrics from~\cite{MartinPredicting2021}, the average log spectral norm had the best Pearson's $r$, which is consistent with the original study.
It is also worth pointing out that \texttt{synflow} outperformed all the metrics from~\cite{MartinPredicting2021}, which might not be that surprising given that it was empirically shown to be performing well in the scenario that requires ranking of ANNs.    
Nevertheless, when this result is compared to the perceptron theory, we observe decent improvement in the quality of predictions in favour of the perceptron theory.  
It should be also noted that the above experiment used networks with different architectures and, e.g., the authors in~\cite{MartinPredicting2021} indicated that their metrics should be used with caution when comparing networks with different architectures.
To us, the results in Table~\ref{tab:complete:anns} are an indicator that to be able to achieve high quality of predictions among  different architectures, the knowledge of statistics of the postsynaptic sums might be an essential information.

\paragraph{Predictions obtained from readout perceptrons}
\label{sect:other:predictions:readout}

To be consistent with the setup of Section~\ref{sect:pred:centroids}, we obtained the statistical metrics when using only the corresponding readout perceptrons (\texttt{synflow} was not considered here). 
These results are presented in Fig.~\ref{fig:weight:watcher:last}; each panel corresponds to a metric. 
Similar to Table~\ref{tab:complete:anns}, Table~\ref{tab:perceptron:anns} reports the overall quality of predictions where the  results for the perceptron theory were obtained according to the methodology described in Section~\ref{sect:pred:centroids}.    
As it could be seen from the table, similar to the previous experiment, the perceptron theory provided much higher quality of predictions. 
Notably, the results of the perceptron theory in Table~\ref{tab:perceptron:anns} are comparable to those of the statistical methods from the previous experiment (cf. Table~\ref{tab:complete:anns}).   
Finally, it is also worth emphasizing that in this experiment's setup, all the methods (including the perceptron theory) formed the predictions without the access to the activations of the last hidden layer (i.e., only a subset of network's weights was used), which suggests that in the case of different architectures the perceptron theory is competitive with the state-of-the-art methods.



\section{Discussion}
\label{sect:discussion}

\subsection{Summary of results} 
\label{sect:discussion:summary}

We present a theory for classification with one-layer perceptrons, which is general in the sense that it applies to networks formed by any learning rule or optimization procedure. 
The curious fact that the theory for perceptrons, the earliest ANN models~\cite{Perceptron}, cannot be found in textbooks might be due to the lack of universality of one-layer perceptrons as a general function approximator~\cite{minkypappert}. 
Here, we trace the late and gradual development of a general perceptron theory, which occurred in the context of neuroscience~\cite{babadi2014sparseness} and more complex neural networks~\cite{babadi2014sparseness, Frady17}.  
The presented perceptron theory generalizes the earlier versions, and, thus, is applicable to a wider variety of neural networks that contain a perceptron-like output layer.

We first verified that our formulation of the perceptron theory Eq.~(\ref{eq:pcorr:mvn}) can accurately predict the performance of echo state networks, even with the readout perceptron optimized by linear regression, the case that could not be treated with the previous formulation of the theory \cite{Frady17}.
Next, we investigated the application of the perceptron theory to shallow and deep networks. 
The empirical evaluation of the predicted accuracy  was performed on numerous classification datasets with shallow networks and on ImageNet with more than a dozen deep neural network architectures. 
In both cases, we observed high correlation between predicted and actual accuracies even when assumptions in the theory definition (Section~\ref{sec:versatile}) were violated, for example, independence of postsynaptic sum distributions for different classes or Gaussianity. 
However, neglecting dependencies introduced a bias to the predicted accuracies (Section~\ref{sect:imagenet:pcorr}), which can be mitigated empirically (Section~\ref{sect:imagenet:bias}). 
Alternatively, we also offer a general formulation in Eq.~(\ref{eq:pcorr:mvn}) that takes the dependencies into account.
One might argue that a theory that predicts accuracies based on the activations in the last hidden layer is not useful, because the accuracies could be computed directly by going through the last layer.   
Note, however, that the theory does not require the activations explicitly, it only requires some low-order statistical moments of the postsynaptic sums in the output neurons, which can be cheaply collected during the execution time. 
In addition, the theory also identifies the low-order statistical moments of the postsynaptic sums as essential for determining the accuracy of the network.

\subsection{Applications and future directions}

We foresee the following applications of this study: 
\noindent
\begin{itemize}
    \item The proposed perceptron theory 
    is applicable to any network architecture with perceptron-like output layer. Thus, the potential application range goes far beyond the network architectures we have investigated in the article.
    
    \item The proposed theory enables ranking of networks with different architectures for a particular task by accessing only low-order statistical moments of the postsynaptic sums in the output neurons, which is important as it avoids a common issue of data privacy.

    \item A more advanced application of our theory is for comparing networks without having any access to the data or the transformation stage of the network. 
    Our experiments showed that the results are less accurate, but maybe still sufficient as the first filtering step in a multi-stage evaluation process for granting access to the data only to the most promising networks. 

    \item Last, our approach can not only predict classification accuracies but also be extended to estimate probabilities of observing a certain vector of postsynaptic sums in the output layer.
    This estimation can potentially be used to detect adversarial examples and other outliers.
    
\end{itemize}

The presented results suggest several novel directions for future investigation. One interesting question is whether the perceptron theory could be used to design more efficient or faster training algorithms, perhaps by using the theoretical predictions
for identifying classes that need more training.  
Another obvious research direction is efficient numerical evaluation of the theory formulae. It would be interesting to compare existing methods for numerical multidimensional integration in their ability to evaluate the more precise formulation in Eq.~(\ref{eq:pcorr:mvn}) for larger number of classes. 



\appendices
\renewcommand\thefigure{S.\arabic{figure}}    
\setcounter{figure}{0} 
\renewcommand\thetable{S.\arabic{table}}    
\setcounter{table}{0} 


\section{Analyzing short-term memory in echo state networks}
\label{sect:capacity:theory}

The theory in~\cite{Frady17} was developed for studying the short-term memory of distributed representations in such models as hyperdimensional computing or, synonymously, vector symbolic architectures (HD/VSA)\footnote{
A comprehensive two-part survey of HD/VSA is available in~\cite{KleykoSurveyVSA2021Part1,KleykoSurveyVSA2021Part2}. 
}
and recurrent randomly connected networks within reservoir computing  (e.g., echo state networks, ESNs~\cite{ESN03}).\footnote{It is worth noting here that HD/VSA and reservoir computing have multiple points of connection. 
There is both a direct connection that has been drawn in, e.g.,~\cite{Frady17,intESN} as well as an implicit connection via cellular automata computations~\cite{YilmazMachine2015,YilmazSymbolic2015,mcdonald2017reservoir} that have also been found useful within HD/VSA~\cite{KleykoBrainsCA2017,SchmuckHardwareOptimizations2019, KleykoCA2020,MenonEmotionHD2021}.  
}
In particular, here we consider the trajectory association task~\cite{PlateBook, HerscheHDMClassifier2021, KleykoDecoding2022} (see below) as one of the ways of studying the short-term memory of a simplified version of the ESN~\cite{intESN}.
The ability to form and use the short-term memory is a key enabler for many HD/VSA use-cases such as representation of data structures~\cite{RachkovskijAnalogical2004,RachkovskijAnalogy2012,OsipovHD_FSA2017, KleykoABF2020, YerxaUCBHD_FSA2018}, processing of strings~\cite{PashchenkoSubstring2020, KleykoPermuted2016, JoshiNgrams2016, RachkovskijRecursiveBinding2022}, and communications~\cite{KleykoMACOM2012, KimHDM2018, WirelessGuirado2022}.
Below, we will introduce the trajectory association task together with the simplified version of the ESN.
We kindly refer readers to~\cite{ESNtut12} for a step-by-step tutorial on applying ESNs; to~\cite{RC09} for a broader overview of the area; and to~\cite{TanakaPhysicalRC2019} for the aspects of physical realization.

The trajectory association task has two stages: memorization and recall.
At the memorization stage, at every timestep $m$  the ESN  stores a symbol $\textbf{s}(m)$ from the sequence of symbols $\textbf{s}$ to be memorized.
The number of unique symbols (i.e., alphabet size) is denoted as $D$.
The symbols are represented using $N$-dimensional random bipolar dense vectors stored in the codebook $\mathbf{\Phi}  \in \{-1,1\}^{N \times D}$.
Thus, at every timestep $m$ the ESN is presented with the corresponding $N$-dimensional vector $\mathbf{\Phi}_{\textbf{s}(m)}$, which is added to the hidden layer of the ESN ($\textbf{x} \in \mathbb{Z}^{N \times 1}$). 
The state of the hidden layer at timestep $m$ (denoted as $\textbf{x}(m)$) is updated as follows: 
\noindent
\begin{equation*}
\textbf{x}(m)= f_\kappa ( \rho(\textbf{x}(m-1)) +  \mathbf{\Phi}_{\textbf{s}(m)}  ),
\label{eq:intESN}
\end{equation*}
\noindent
where $\textbf{x}(m-1)$ is the previous state of the hidden layer at timestep $m-1$;
$\rho$ denotes the permutation operation (e.g., circular shift to the right), which acts as a simple variant of a recurrent connectivity matrix;
$f_\kappa (x)$ is a clipping function -- nonlinear activation function, which keeps the values of the hidden layer in the limited range using  a threshold value $\kappa$ as: 
\noindent
\begin{equation*}
f_\kappa (x) = 
\begin{cases}
-\kappa & x \leq -\kappa \\
x & -\kappa < x < \kappa. \\
\kappa & x \geq \kappa
\end{cases}
\label{eq:clipping}
\end{equation*}
\noindent
In practice, the value of $\kappa$  regulates the recency effect of the ESN.

At the recall stage, the ESN uses the content of its hidden layer $\textbf{x}(m)$ as the query vector to retrieve the symbol stored $d$ steps ago, where $d$ denotes delay. 
In the experiments below, the range of the delays varied between $0$ and $25$.
The recall is done by using the readout perceptron for particular $d$, which contains one $N$-dimensional vector (linear filter) per each symbol.
The readout perceptron is denoted as $\textbf{W}^{d} \in \mathbb{R}^{D \times N}$ and the recall is done as: 
\noindent
\begin{equation*}
\hat{\textbf{s}}(m-d)=\argmax ( \textbf{W}^{d} \textbf{x}(m)  ),
\label{eq:recall}
\end{equation*}
\noindent
where $\argmax(\cdot)$ returns the symbol with the highest postsynaptic sum among the output neurons for the chosen delay $\textbf{W}^{d}$ value and the given hidden layer state $\textbf{x}(m)$.

Let us consider two approaches of forming the readout perceptron. 
First, it can be constructed as done usually in HD/VSA from the codebook $\mathbf{\Phi}$ and the reverse permutation by $d$:
\noindent
\begin{equation}
\textbf{W}^{d}=\rho^{-d}(\mathbf{\Phi}^\top).
\label{eq:readoutVSAs}
\end{equation}
\noindent
The advantage of this approach is that no training is required to obtain the readout perceptron. 

An alternative approach, which is more native for ESNs, is to obtain $\textbf{W}^{d}$ via solving a linear regression on a given training sequence and the corresponding states of the hidden layer. 
The advantage of this approach is that, as we have seen in Fig.~\ref{fig:ESN:acc} in the main text, it has higher accuracy, than the codebook-based approach.

\begin{figure}[tb]
\centering
\includegraphics[width=1.0\linewidth]{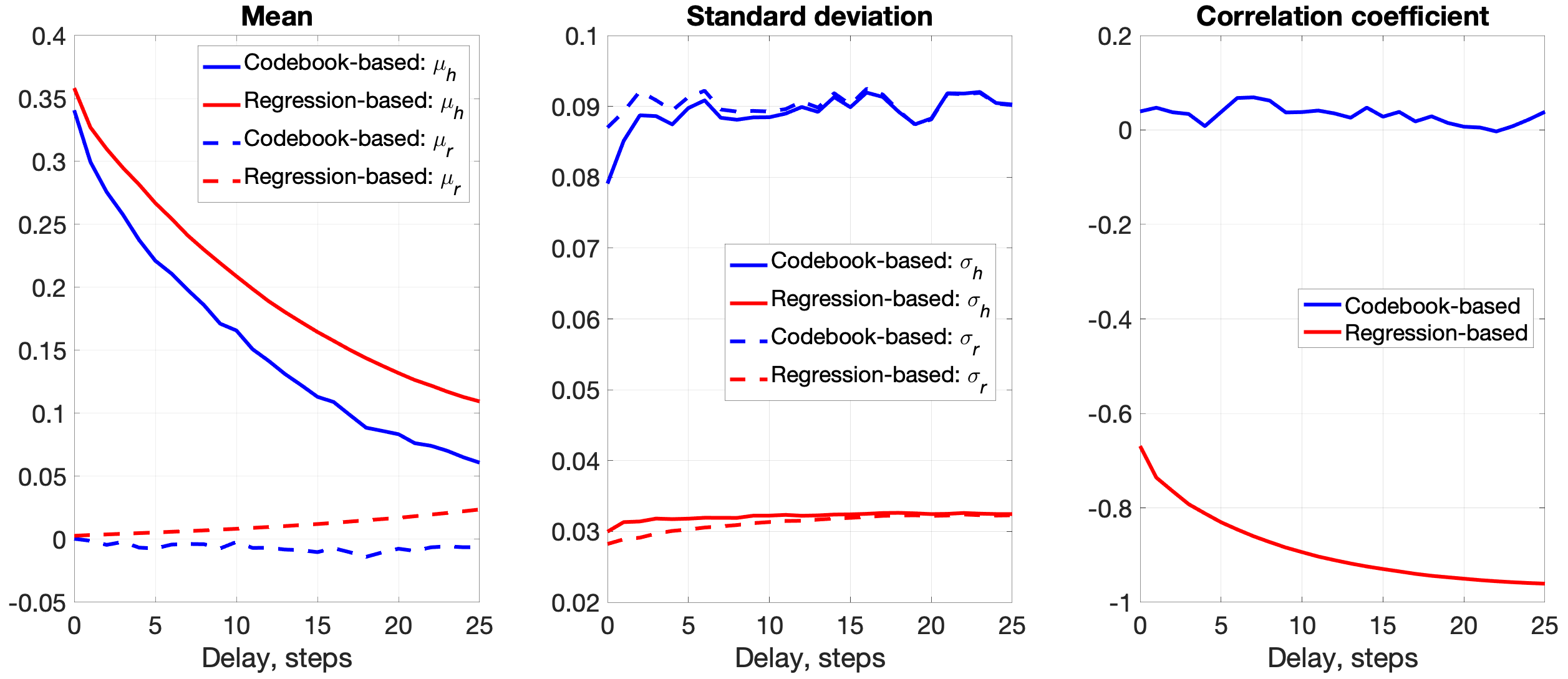}
\caption{
The statistics extracted for the case of codebook-based and regression-based perceptrons. 
The following values for the ESN parameters were used: $N=100$, $D=2$, $\kappa=4$.
The length of test sequences was $10,000$. 
All reported valued were averaged over $50$ simulations with random codebooks. 
}
\label{fig:ESN:stat}
\end{figure}

While it is straightforward to simulate the presented network and obtain the accuracies for different values of $d$ empirically, predicting the accuracy analytically given only the parameters ($N$, $D$, $d$, and $\kappa$) is challenging. 
Nevertheless,~\cite{Frady17} has proposed a solution to this problem when the perceptron is based on the codebook as in Eq.~(\ref{eq:readoutVSAs}). 
The solution includes two components. 
The first one is the equation for calculating the predicted accuracies $a$ (i.e., the expected accuracy):
\noindent
\begin{equation}
\begin{split}
&a := p(\textbf{s}(m-d) = \hat{\textbf{s}}(m-d))= \\ &\int_{-\infty}^{\infty} \frac{dx}{\sqrt{2 \pi} \sigma_{h}} e^{-\frac{ (x-\mu_{h})^2 }{2 \sigma_{h}^2}} ( \Phi(x,\mu_{r}, \sigma_{r}))^{D-1},
\end{split}
 \label{eq:pcorr:orig}
 \end{equation}
\noindent
where $\mu_{h}$ and $\sigma_{h}$ denote the mean and standard deviation of the postsynaptic sum (i.e., dot product) between $\textbf{x}(m)$ and the row of $\textbf{W}^{d}$ (i.e., linear filter) corresponding to the correct symbol $\textbf{s}(m-d)$ while $\mu_{r}$ and $\sigma_{r}$ denote the mean and standard deviation of the postsynaptic sum for all other symbols in the codebook. 
Note that Eq.~(\ref{eq:pcorr:orig}) is equivalent to Eq.~(\ref{eq:pcorr:orig1}) in the main text. 
Moreover, Eq.~(\ref{eq:pcorr:orig}) is a special case of Eq.~(\ref{eq:pcorr}) where for all symbols in the codebook other than the correct symbol the same $\mu_{r}$ and $\sigma_{r}$ are assumed that is because the codebook-based perceptron uses $\mathbf{\Phi}$ where representations for different symbols are random and, therefore, for large $N$ they are quasi orthogonal to each other. 
Due to this fact, we know that the expected value of $\mu_{r}$ is $0$ and that of $\sigma_{r}$ is $\sqrt{N\kappa(\kappa+1)/3}$. 
However, the values of $\mu_{h}$ and $\sigma_{h}$ depend on the given values of $d$ and $\kappa$, therefore, the second component of the solution determines them (please refer to Eqs.~(2.44)-(2.47) in~\cite{Frady17}).

It is also worth mentioning that Eq.~(\ref{eq:pcorr:mvn}) is the generalization of Eq.~(\ref{eq:pcorr}) because once we assume that $\bm{\Sigma}$ is diagonal (i.e., variables in $\textbf{x}$ are independent) Eq.~(\ref{eq:pcorr:mvn}) can be simplified to Eq.~(\ref{eq:pcorr}) as follows:  
\noindent
\begin{equation*}
\begin{split}
&\textbf{a}_i= 
\int_{-\infty}^{\infty} 
\int_{-\infty}^{\textbf{x}_1} 
\dots 
\int_{-\infty}^{\textbf{x}_1} 
p(\textbf{x}, \bm{\mu}, \bm{\Sigma}) d\textbf{x}_L \dots d\textbf{x}_1 = \\
&\int_{-\infty}^{\infty} 
p(\textbf{x}_1)  d\textbf{x}_1 
\int_{-\infty}^{\textbf{x}_1} 
p(\textbf{x}_2)  d\textbf{x}_2
\dots 
\int_{-\infty}^{\textbf{x}_1} 
p(\textbf{x}_L) d\textbf{x}_L = \\
&
\int_{-\infty}^{\infty} 
p(\textbf{x}_1,\bm{\mu}_{1},\bm{\sigma}_{1})  d\textbf{x}_1 
\Phi(\textbf{x}_1,\bm{\mu}_{2}, \bm{\sigma}_{2})
\dots 
\Phi(\textbf{x}_1,\bm{\mu}_{L}, \bm{\sigma}_{L})
= \\
&\int_{-\infty}^{\infty} 
p(\textbf{x}_1,\bm{\mu}_{1},\bm{\sigma}_{1})  d\textbf{x}_1 
\prod_{j=2}^{L}
\Phi(\textbf{x}_1,\bm{\mu}_{j}, \bm{\sigma}_{j}).
\end{split}
\label{eq:mvn:to:pcorr}
 \end{equation*}
\noindent

\begin{figure}[tb]
\centering
\includegraphics[width=1.0\columnwidth]{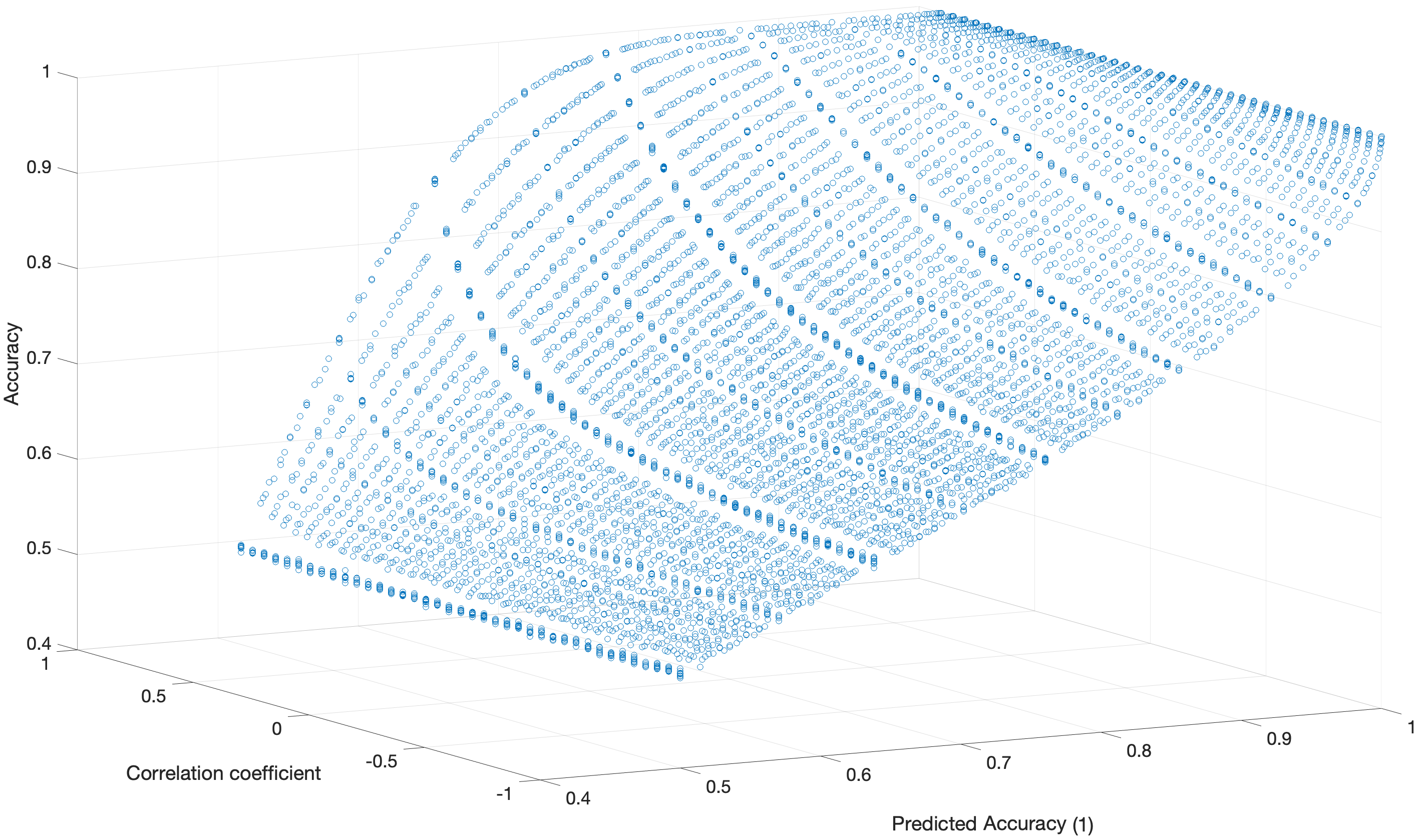}
\caption{
The actual accuracy of a synthetic datasets for different Pearson correlation coefficients and predicted accuracy calculated assuming that the correlation between classes is zero, i.e., according to Eqs.~(\ref{eq:pcorr:orig1})/(\ref{eq:pcorr})/(\ref{eq:pcorr:orig}), which are all the same for $D=2$.
}
\label{fig:2D:pcorr}
\end{figure}

Note that in the case of the regression-based perceptron one cannot assume the independence of linear filters in the perceptron. 
Therefore, currently, there is no way of analytically estimating $\mu_{h}$, $\mu_{r}$, $\sigma_{h}$, and $\sigma_{r}$ for the regression-based perceptron. 
Nevertheless, these values could be estimated empirically using the simulations. 
Fig.~\ref{fig:ESN:stat} presents the $\mu_{h}$, $\mu_{r}$, $\sigma_{h}$, and $\sigma_{r}$ for different values of $d$ for both regression-based (red color) and codebook-based (blue color) perceptrons.\footnote{Since the regression-based and codebook-based perceptrons might have different norms Fig.~\ref{fig:ESN:stat} was obtained using cosine similarities instead of postsynaptic sums. 
Obviously, the usage of the cosine similarity  does not affect neither the accuracy nor the applicability of Eq.~(\ref{eq:pcorr:orig}) to the estimated statistics.
}
Notice, that there are several important differences between the statistics observed for different perceptrons. 
First, $\mu_{h}$ for the regression-based perceptron is higher than that of the codebook-based perceptron, however, for both perceptrons   $\mu_{h}$ is decreasing with the increased delay, which is expected.
Second, both $\sigma_{h}$, and $\sigma_{r}$ are much lower in the case of the regression-based perceptron. 
Both facts should positively affect the accuracy. 
Third, there is a strong negative correlation between the linear filters in the regression-based perceptron.
As it was shown in Fig.~\ref{fig:ESN:acc}  in the main text, the presence of correlation hindered the applicability of Eq.~(\ref{eq:pcorr:orig}) but the use of Eq.~(\ref{eq:pcorr:mvn}) allowed getting the correct accuracy.

Last, let us explain how the original problem: analysis of trajectory association with ESN relates to predicting the accuracy of a neural network. First, in the case of classification we do not assume delay so we can safely say that $d=0$.
Second, the activation of the last hidden layer of an neural network corresponds to the hidden layer activity in the considered ESN.  
Note, however, that there is no guarantee that these activations are noisy versions of the entries of some fixed random matrix as the activations  come from (usually continuous) real  data.
Last, the weights of the output layer of a neural network are more similar to the regression-based perceptron than to the codebook-based perceptron because they are obtained through an optimization procedure (e.g., error backprogation). 
This implicitly explains the bias observed in Fig.~\ref{fig:in:pcorr:norm} when using Eq.~(\ref{eq:pcorr}) to calculate the predicted accuracy. 
Recall, that in Fig.~\ref{fig:in:pcorr:norm} we have observed both underestimation and overestimation of the empirical accuracy while in Fig.~\ref{fig:ESN:acc} we have observed only the overestimation. 
Therefore, in Section~\ref{sect:pred:corr} we will study the effect of the correlation on the results produced by Eq.~(\ref{eq:pcorr}) or Eqs.~(\ref{eq:pcorr:orig1})/(\ref{eq:pcorr:orig}).

\section{The effect of correlations} 
\label{sect:pred:corr}

In Fig.~\ref{fig:ESN:acc} in the main text, we have seen that even in the case of two symbols (or two classes) the presence of correlation might result in inaccuracies between the predicted accuracy and the actual accuracy.  
Nevertheless, using Eqs.~(\ref{eq:pcorr:orig1})/(\ref{eq:pcorr:orig}) or Eq.~(\ref{eq:pcorr}) to calculate the predicted accuracy is tempting because numerically it is a simple task compared to Eq.~(\ref{eq:pcorr:mvn}). 
Therefore, one interesting question is whether it is possible to compensate bias introduced by Eqs.~(\ref{eq:pcorr:orig1})/(\ref{eq:pcorr:orig}) or Eq.~(\ref{eq:pcorr}) without the need of, e.g., calculating the accuracies on sub-problems as in Fig.~\ref{fig:in:pcorr:norm:sub}.

In order to study this, we created a synthetic binary classification problem ($D=2$) assuming that the mean value of postsynaptic sums of the incorrect class is $0$ while that of the correct class is a parameter. 
It was also assumed that postsynaptic sums distributions of both classes have the same standard deviation, which is also a parameter. 
Finally, the  postsynaptic sum distributions can correlate with each other positively or negatively and their Pearson correlation coefficient is a parameter. 
Note that when there is no correlation between the classes we can perfectly use Eqs.~(\ref{eq:pcorr:orig1})/(\ref{eq:pcorr:orig}) or Eq.~(\ref{eq:pcorr}) for the given mean and standard deviation values. 
We can also use Eqs.~(\ref{eq:pcorr:orig1})/(\ref{eq:pcorr:orig}) or Eq.~(\ref{eq:pcorr}) even when there is a correlation but in this case the predicted accuracy would differ from the actual accuracy. 
In order to see the effect of the correlation between classes on the accuracy we simulated many cases for different values of correlation, mean, and standard deviation.
Fig.~\ref{fig:2D:pcorr} depicts the results. 
As we can see, there is a highly non-linear relation between the predicted accuracy, Pearson correlation coefficient, and the empirical accuracy. 
Qualitatively, we expect that in the case of a negative correlation the accuracy will be lower than the predicted accuracies (cf. red lines in Fig.~\ref{fig:ESN:acc}) and vice versa in the case of the positive correlation.
However, it seems hard to make more quantitative correction without making a look-up table like structure, which would interpolate the expected accuracy value for the given correlation and predicted accuracy. 
While this is possible to do for the binary classification problems, this solution becomes inadequate for larger values of $D$ because we would have to deal with many interactions in the covariance matrix, which cannot be simply stored as a compact look-up table.

\label{sect:bib}
\bibliographystyle{unsrt}
\bibliography{bica}

\begin{thebibliography}{100}

\bibitem{HLAIEthics}
N.~A. Smuha.
\newblock {The EU Approach to Ethics Guidelines for Trustworthy Artificial
  Intelligence}.
\newblock {\em {Computer Law Review International}}, 20(4):97--106, 2019.

\bibitem{USAI16}
{Executive Office of the President National Science and Technology Council
  Committee on Technology}.
\newblock {\em {Preparing for the Future of Artificial Intelligence}}.
\newblock 2016.

\bibitem{papyan2020prevalence}
V.~Papyan, X.~Y. Han, and D.~L. Donoho.
\newblock {Prevalence of Neural Collapse During the Terminal Phase of Deep
  Learning Training}.
\newblock {\em Proceedings of the National Academy of Sciences},
  117(40):24652--24663, 2020.

\bibitem{GeirhosShortcut2020}
R.~Geirhos, J.-H. Jacobsen, C.~Michaelis, R.~Zemel, W.~Brendel, M.~Bethge, and
  F.~A. Wichmann.
\newblock {Shortcut Learning in Deep Neural Networks}.
\newblock {\em {Nature Machine Intelligence}}, 2(11):665--673, 2020.

\bibitem{shwartz2017opening}
R.~Shwartz-Ziv and N.~Tishby.
\newblock {Opening the Black Box of Deep Neural Networks via Information}.
\newblock {\em {arXiv:1703.00810}}, pages 1--19, 2017.

\bibitem{BommasaniOpportunities2021}
R.~Bommasani, D.~A. Hudson, E.~Adeli, R.~Altman, S.~Arora, S.~von Arx, M.~S.
  Bernstein, J.~Bohg, A.~Bosselut, E.~Brunskill, et~al.
\newblock {On the Opportunities and Risks of Foundation Models}.
\newblock {\em arXiv:2108.07258}, pages 1--214, 2021.

\bibitem{egmont1994quality}
M.~Egmont-Petersen, J.~L. Talmon, J.~Brender, and P.~McNair.
\newblock {On the Quality of Neural Net Classifiers}.
\newblock {\em Artificial Intelligence in Medicine}, 6(5):359--381, 1994.

\bibitem{UnterthinerPredAcc2020}
T.~Unterthiner, D.~Keysers, S.~Gelly, O.~Bousquet, and I.~Tolstikhin.
\newblock {Predicting Neural Network Accuracy from Weights}.
\newblock {\em arXiv:2002.11448}, pages 1--18, 2020.

\bibitem{DeChantPredAcc2019}
C.~DeChant, S.~Han, and H.~Lipson.
\newblock {Predicting the Accuracy of Neural Networks from Final and
  Intermediate Layer Outputs}.
\newblock In {\em {International Conference on Learning Representations
  (ICLR)}}, pages 1--6, 2019.

\bibitem{Jaeger12}
H.~Jaeger.
\newblock {Long Short-Term Memory in Echo State Networks: Details of a
  Simulation Study}.
\newblock Technical report, (Technical Report 27). Bremen: Jacobs Univesity,
  2012.

\bibitem{Frady17}
E.~P. Frady, D.~Kleyko, and F.~T. Sommer.
\newblock {A Theory of Sequence Indexing and Working Memory in Recurrent Neural
  Networks}.
\newblock {\em Neural Computation}, 30:1449--1513, 2018.

\bibitem{feraud2002methodology}
R.~F{\'e}raud and F.~Cl{\'e}rot.
\newblock {A Methodology to Explain Neural Network Classification}.
\newblock {\em Neural Networks}, 15(2):237--246, 2002.

\bibitem{baehrens2010explain}
D.~Baehrens, T.~Schroeter, S.~Harmeling, M.~Kawanabe, K.~Hansen, and K.-R.
  Muller.
\newblock {How to Explain Individual Classification Decisions}.
\newblock {\em Journal of Machine Learning Research}, 11:1803--1831, 2010.

\bibitem{deng2017peephole}
B.~Deng, J.~Yan, and D.~Lin.
\newblock {Peephole: Predicting Network Performance before Training}.
\newblock {\em arXiv:1712.03351}, pages 1--10, 2017.

\bibitem{istrate2019tapas}
R.~Istrate, F.~Scheidegger, G.~Mariani, D.~Nikolopoulos, C.~Bekas, and A.~C.~I.
  Malossi.
\newblock {TAPAS: Train-less Accuracy Predictor for Architecture Search}.
\newblock In {\em Thirty-Third AAAI Conference on Artificial Intelligence
  (AAAI)}, volume~33, pages 3927--3934, 2019.

\bibitem{elsken2018neural}
T.~Elsken, J.~H. Metzen, and F.~Hutter.
\newblock {Neural Architecture Search: A Survey}.
\newblock {\em Journal of Machine Learning Research}, 20(55):1--21, 2019.

\bibitem{baker2017accelerating}
B.~Baker, O.~Gupta, R.~Raskar, and N.~Naik.
\newblock {Accelerating Neural Architecture Search Using Performance
  Prediction}.
\newblock In {\em {International Conference on Learning Representations
  (ICLR)}}, pages 1--19, 2018.

\bibitem{romero2014fitnets}
A.~Romero, N.~Ballas, S.~E. Kahou, A.~Chassang, C.~Gatta, and Y.~Bengio.
\newblock {FitNets: Hints for Thin Deep Nets}.
\newblock {\em arXiv:1412.6550}, pages 1--13, 2014.

\bibitem{dhurandhar2018improving}
A.~Dhurandhar, K.~Shanmugam, R.~Luss, and P.~A. Olsen.
\newblock {Improving Simple Models with Confidence Profiles}.
\newblock In {\em Advances in Neural Information Processing Systems (NeurIPS)},
  pages 10296--10306, 2018.

\bibitem{MONTAVON20181}
G.~Montavon, W.~Samek, and K.-R. Müller.
\newblock {Methods for Interpreting and Understanding Deep Neural Networks}.
\newblock {\em {Digital Signal Processing}}, 73:1--15, 2018.

\bibitem{Zeiler14}
M.~D. Zeiler and R.~Fergus.
\newblock {Visualizing and Understanding Convolutional Networks}.
\newblock In {\em {European Conference on Computer Vision (ECCV)}}, pages
  818--833, 2014.

\bibitem{MartinPredicting2021}
C.~H. Martin, T.~S. Peng, and M.~W. Mahoney.
\newblock {Predicting Trends in the Quality of State-of-the-art Neural Networks
  without Access to Training or Testing Data}.
\newblock {\em {Nature Communications}}, 12(1):1--13, 2021.

\bibitem{FixClassifier}
E.~Hoffer, I.~Hubara, and D.~Soudry.
\newblock {Fix Your Classifier: the Marginal Value of Training the Last Weight
  Layer}.
\newblock In {\em {International Conference on Learning Representations
  (ICLR)}}, pages 1--11, 2018.

\bibitem{ZhouEconas2020}
D.~Zhou, X.~Zhou, W.~Zhang, C.~C. Loy, S.~Yi, X.~Zhang, and W.~Ouyang.
\newblock {EcoNAS: Finding Proxies for Economical Neural Architecture Search}.
\newblock In {\em {IEEE Conference on Computer Vision and Pattern Recognition
  (CVPR)}}, pages 11396--11404, 2020.

\bibitem{AbdelfattahZero2021}
M.~S. Abdelfattah, A.~Mehrotra, L.~Dudziak, and N.~D. Lane.
\newblock {Zero-cost Proxies for Lightweight NAS}.
\newblock In {\em {International Conference on Learning Representations
  (ICLR)}}, pages 1--17, 2021.

\bibitem{MellorNeural2021}
J.~Mellor, J.~Turner, A.~Storkey, and E.~J. Crowley.
\newblock {Neural Architecture Search without Training}.
\newblock In {\em {International Conference on Machine Learning (ICML)}}, pages
  7588--7598, 2021.

\bibitem{LeeSNIP2019}
N.~Lee, T.~Ajanthan, and P.~H.~S. Torr.
\newblock {SNIP: Single-shot Network Pruning based on Connection Sensitivity}.
\newblock In {\em {International Conference on Learning Representations
  (ICLR)}}, pages 1--15, 2019.

\bibitem{TurnerBlockswap2019}
J.~Turner, E.~J. Crowley, M.~O'Boyle, A.~Storkey, and G.~Gray.
\newblock {BlockSwap: Fisher-guided Block Substitution for Network Compression
  on a Budget}.
\newblock In {\em {International Conference on Learning Representations
  (ICLR)}}, pages 1--15, 2020.

\bibitem{WangWinningTicket2020}
C.~Wang, G.~Zhang, and R.~Grosse.
\newblock {Picking Winning Tickets before Training by Preserving Gradient
  Flow}.
\newblock In {\em {International Conference on Learning Representations
  (ICLR)}}, pages 1--11, 2020.

\bibitem{tanaka2020pruning}
H.~Tanaka, D.~Kunin, D.~L. Yamins, and S.~Ganguli.
\newblock {Pruning Neural Networks without Any Data by Iteratively Conserving
  Synaptic Flow}.
\newblock In {\em {Advances in Neural Information Processing Systems
  (NeurIPS)}}, volume~33, pages 6377--6389, 2020.

\bibitem{Perceptron}
F.~Rosenblatt.
\newblock {The Perceptron — a Perceiving and Recognizing Automaton}.
\newblock Technical report, Report 85–460–1, Cornell Aeronautical
  Laboratory, 1957.

\bibitem{rosenblatt1961principles}
F.~Rosenblatt.
\newblock {Principles of Neurodynamics. Perceptrons and the Theory of Brain
  Mechanisms}.
\newblock Technical report, {Cornell Aeronautical Lab Inc Buffalo, NY}, 1961.

\bibitem{steinbuch1963learning}
K.~Steinbuch and U.~W. Piske.
\newblock {Learning Matrices and Their Applications}.
\newblock {\em {IEEE Transactions on Electronic Computers}}, EC-12(6):846--862,
  1963.

\bibitem{amari1977competition}
S.~Amari and M.~A. Arbib.
\newblock {Competition and Cooperation in Neural Nets}.
\newblock {\em {Systems Neuroscience}}, pages 119--165, 1977.

\bibitem{grossberg1988nonlinear}
S.~Grossberg.
\newblock {Nonlinear Neural Networks: Principles, Mechanisms, and
  Architectures}.
\newblock {\em {Neural Networks}}, 1(1):17--61, 1988.

\bibitem{maass2000computational}
W.~Maass.
\newblock {On the Computational Power of Winner-Take-All}.
\newblock {\em {Neural Computation}}, 12(11):2519--2535, 2000.

\bibitem{PetersonEtAl1954}
W.~Peterson, T.~Birdsall, and W.~Fox.
\newblock {The Theory of Signal Detectability}.
\newblock {\em Proceedings of the IRE Professional Group on Information
  Theory}, 4:171--212, 1954.

\bibitem{minkypappert}
M.~Minsky and S.~A. Papert.
\newblock {\em {Perceptrons: An Introduction to Computational Geometry}}.
\newblock The MIT Press, 1969.

\bibitem{babadi2014sparseness}
B.~Babadi and H.~Sompolinsky.
\newblock {Sparseness and Expansion in Sensory Representations}.
\newblock {\em Neuron}, 83(5):1213--1226, 2014.

\bibitem{Kanerva09}
P.~Kanerva.
\newblock {Hyperdimensional Computing: An Introduction to Computing in
  Distributed Representation with High-Dimensional Random Vectors}.
\newblock {\em Cognitive Computation}, 1(2):139--159, 2009.

\bibitem{PlateBook}
T.~A. Plate.
\newblock {\em {Holographic Reduced Representations: Distributed Representation
  for Cognitive Structures}}.
\newblock {Stanford: Center for the Study of Language and Information (CSLI)},
  2003.

\bibitem{KleykoSurveyVSA2021Part1}
D.~Kleyko, D.~A. Rachkovskij, E.~Osipov, and A.~Rahimi.
\newblock {A Survey on Hyperdimensional Computing aka Vector Symbolic
  Architectures, Part I: Models and Data Transformations}.
\newblock {\em {ACM Computing Surveys}}, 55(6):1--40, 2022.

\bibitem{KleykoSurveyVSA2021Part2}
D.~Kleyko, D.~A. Rachkovskij, E.~Osipov, and A.~Rahimi.
\newblock {A Survey on Hyperdimensional Computing aka Vector Symbolic
  Architectures, Part II: Applications, Cognitive Models, and Challenges}.
\newblock {\em {ACM Computing Surveys}}, 2022.

\bibitem{KleykoComputingParadigm2021}
D.~Kleyko, M.~Davies, E.~P. Frady, P.~Kanerva, S.~J. Kent, B.~A. Olshausen,
  E.~Osipov, J.~M. Rabaey, D.~A. Rachkovskij, A.~Rahimi, and F.~T. Sommer.
\newblock {Vector Symbolic Architectures as a Computing Framework for Emerging
  Hardware}.
\newblock {\em {Proceedings of the IEEE}}, 110(10):1538--1571, 2022.

\bibitem{Scalarencoding}
D.~A. Rachkovskij, S.~V. Slipchenko, E.~M. Kussul, and T.~N. Baidyk.
\newblock {Sparse Binary Distributed Encoding of Scalars}.
\newblock {\em Journal of Automation and Information Sciences}, 37(6):12--23,
  2005.

\bibitem{frady2021computing}
E.~P. Frady, D.~Kleyko, C.~J. Kymn, B.~A. Olshausen, and F.~T. Sommer.
\newblock {Computing on Functions Using Randomized Vector Representations}.
\newblock {\em {arXiv:2109.03429}}, pages 1--33, 2021.

\bibitem{FradyFunctionsNICE2022}
E.~P. Frady, D.~Kleyko, C.~J. Kymn, B.~A. Olshausen, and F.~T. Sommer.
\newblock {Computing on Functions Using Randomized Vector Representations (in
  brief)}.
\newblock In {\em {Neuro-Inspired Computational Elements Conference (NICE)}},
  pages 115--122, 2022.

\bibitem{HDGestureIEEE}
A.~Rahimi, P.~Kanerva, L.~Benini, and J.~M. Rabaey.
\newblock {Efficient Biosignal Processing Using Hyperdimensional Computing:
  Network Templates for Combined Learning and Classification of ExG Signals}.
\newblock {\em Proceedings of the IEEE}, 107(1):123--143, 2019.

\bibitem{KleykoTradeoffs2018}
D.~Kleyko, A.~Rahimi, D.~A. Rachkovskij, E.~Osipov, and J.~M. Rabaey.
\newblock {Classification and Recall with Binary Hyperdimensional Computing:
  Tradeoffs in Choice of Density and Mapping Characteristic}.
\newblock {\em IEEE Transactions on Neural Networks and Learning Systems},
  29(12):5880--5898, 2018.

\bibitem{GeClassificationReview2020}
L.~Ge and K.~K. Parhi.
\newblock {Classification using Hyperdimensional Computing: A Review}.
\newblock {\em IEEE Circuits and Systems Magazine}, 20(2):30--47, 2020.

\bibitem{PlateTr}
T.~A. Plate.
\newblock {Holographic Reduced Representations}.
\newblock {\em IEEE Transactions on Neural Networks}, 6(3):623--641, 1995.

\bibitem{Owen1980}
D.~B. Owen.
\newblock {A Table of Normal Integrals}.
\newblock {\em Communications in Statistics: Simulation and Computation},
  9(4):389--419, 1980.

\bibitem{HDNP17}
A.~Rahimi, S.~Datta, D.~Kleyko, E.~P. Frady, B.~Olshausen, P.~Kanerva, and
  J.~M. Rabaey.
\newblock {High-dimensional Computing as a Nanoscalable Paradigm}.
\newblock {\em Circuits and Systems I: Regular Papers, IEEE Transactions on},
  64(9):2508--2521, 2017.

\bibitem{Gallant13}
S.~I. Gallant and T.~W. Okaywe.
\newblock {Representing Objects, Relations, and Sequences}.
\newblock {\em Neural Computation}, 25(8):2038--2078, 2013.

\bibitem{KleykoHolographic2017}
D.~Kleyko, E.~Osipov, A.~Senior, A.~I. Khan, and Y.~A. Sekercioglu.
\newblock {Holographic Graph Neuron: A Bio-inspired Architecture for Pattern
  Processing}.
\newblock {\em IEEE Transactions on Neural Networks and Learning Systems},
  28(6):1250--1262, 2017.

\bibitem{Rachkovskij2001}
D.~A. Rachkovskij.
\newblock {Representation and Processing of Structures with Binary Sparse
  Distributed Codes}.
\newblock {\em IEEE Transactions on Knowledge and Data Engineering},
  3(2):261--276, 2001.

\bibitem{FradySDR2020}
E.~P. Frady, D.~Kleyko, and F.~T. Sommer.
\newblock {Variable Binding for Sparse Distributed Representations: Theory and
  Applications}.
\newblock {\em {IEEE Transactions on Neural Networks and Learning Systems}},
  99(PP):1--14, 2021.

\bibitem{ESN03}
H.~Jaeger.
\newblock {Adaptive Nonlinear System Identification with Echo State Networks}.
\newblock In {\em {Advances in Neural Information Processing Systems
  (NeurIPS)}}, pages 593--600, 2003.

\bibitem{ESNtut12}
M.~Lukosevicius.
\newblock {A Practical Guide to Applying Echo State Networks}.
\newblock In {\em {Neural Networks: Tricks of the Trade}}, volume 7700 of {\em
  Lecture Notes in Computer Science}, pages 659--686, 2012.

\bibitem{RVFLorig}
B.~Igelnik and Y.~H. Pao.
\newblock {Stochastic Choice of Basis Functions in Adaptive Function
  Approximation and the Functional-Link Net}.
\newblock {\em IEEE Transactions on Neural Networks}, 6:1320--1329, 1995.

\bibitem{Dua:2019}
D.~Dua and C.~Graff.
\newblock {UCI Machine Learning Repository}, 2019.

\bibitem{HundredsClassifiers}
M.~Fernandez-Delgado, E.~Cernadas, S.~Barro, and D.~Amorim.
\newblock {Do we Need Hundreds of Classifiers to Solve Real World
  Classification Problems?}
\newblock {\em Journal of Machine Learning Research}, 15:3133--3181, 2014.

\bibitem{intRVFL}
D.~Kleyko, M.~Kheffache, E.~P. Frady, U.~Wiklund, and E.~Osipov.
\newblock {Density Encoding Enables Resource-Efficient Randomly Connected
  Neural Networks}.
\newblock {\em {IEEE Transactions on Neural Networks and Learning Systems}},
  32(8):3777--3783, 2021.

\bibitem{KussulDiagnostics1998}
E.~M. Kussul, L.~M. Kasatkina, D.~A. Rachkovskij, and D.~C. Wunsch.
\newblock {Application of Random Threshold Neural Networks for Diagnostics of
  Micro Machine Tool Condition}.
\newblock In {\em International Joint Conference on Neural Networks (IJCNN)},
  volume~1, pages 241--244, 1998.

\bibitem{goltsev2005combination}
A.~Goltsev and D.~A. Rachkovskij.
\newblock {Combination of the Assembly Neural Network with a Perceptron for
  Recognition of Handwritten Digits Arranged in Numeral Strings}.
\newblock {\em {Pattern Recognition}}, 38(3):315--322, 2005.

\bibitem{Rasanen2015tr}
O.~Rasanen and J.~Saarinen.
\newblock {Sequence Prediction with Sparse Distributed Hyperdimensional Coding
  Applied to the Analysis of Mobile Phone Use Patterns}.
\newblock {\em IEEE Transactions on Neural Networks and Learning Systems},
  27(9):1878--1889, 2016.

\bibitem{BICA16}
D.~Kleyko, E.~Osipov, and D.A. Rachkovskij.
\newblock {Modification of Holographic Graph Neuron using Sparse Distributed
  Representations}.
\newblock {\em {Procedia Computer Science}}, 88:39--45, 2016.

\bibitem{KleykoIndustrial2018}
D.~Kleyko, E.~Osipov, N.~Papakonstantinou, and V.~Vyatkin.
\newblock {Hyperdimensional Computing in Industrial Systems: The Use-Case of
  Distributed Fault Isolation in a Power Plant}.
\newblock {\em IEEE Access}, 6:30766--30777, 2018.

\bibitem{DiaoGLVQHD2021}
C.~Diao, D.~Kleyko, J.~M. Rabaey, and B.~A. Olshausen.
\newblock {Generalized Learning Vector Quantization for Classification in
  Randomized Neural Networks and Hyperdimensional Computing}.
\newblock In {\em {International Joint Conference on Neural Networks (IJCNN)}},
  pages 1--9, 2021.

\bibitem{HuangSpeaker2022}
P.-C. Huang, D.~Kleyko, J.~M. Rabaey, B.~A. Olshausen, and P.~Kanerva.
\newblock {Computing with Hypervectors for Efficient Speaker Identification}.
\newblock {\em {arXiv:2208.13285}}, pages 1--5, 2022.

\bibitem{ImageNet}
O.~Russakovsky, J.~Deng, H.~Su, J.~Krause, S.~Satheesh, S.~Ma, Z.~Huang,
  A.~Karpathy, A.~Khosla, M.~Bernstein, A.~C. Berg, and L.~Fei-Fei.
\newblock {ImageNet Large Scale Visual Recognition Challenge}.
\newblock {\em International Journal of Computer Vision}, 115:211--252, 2015.

\bibitem{AlexNet}
A.~Krizhevsky, I.~Sutskever, and G.~E. Hinton.
\newblock {ImageNet Classification with Deep Convolutional Neural Networks}.
\newblock In {\em {Advances in Neural Information Processing Systems
  (NeurIPS)}}, pages 1097--1105, 2012.

\bibitem{GoogLeNet}
C.~Szegedy, W.~Liu, Y.~Jia, P.~Sermanet, S.~Reed, D.~Anguelov, D.~Erhan,
  V.~Vanhoucke, and A.~Rabinovich.
\newblock {Going Deeper with Convolutions}.
\newblock In {\em {IEEE Conference on Computer Vision and Pattern Recognition
  (CVPR)}}, pages 1--9, 2015.

\bibitem{ResNet}
K.~He, X.~Zhang, S.~Ren, and J.~Sun.
\newblock {Deep Residual Learning for Image Recognition}.
\newblock In {\em {IEEE Conference on Computer Vision and Pattern Recognition
  (CVPR)}}, pages 770--778, 2016.

\bibitem{VGG}
K.~Simonyan and A.~Zisserman.
\newblock {Very Deep Convolutional Networks for Large-Scale Image Recognition}.
\newblock {\em arXiv:1409.1556}, pages 1--14, 2014.

\bibitem{ShuffleNet}
X.~Zhang, X.~Zhou, M.~Lin, and J.~Sun.
\newblock {ShuffleNet: An Extremely Efficient Convolutional Neural Network for
  Mobile Devices}.
\newblock In {\em {IEEE Conference on Computer Vision and Pattern Recognition
  (CVPR)}}, pages 6848--6856, 2018.

\bibitem{MobileNet}
M.~Sandler, A.~Howard, M.~Zhu, A.~Zhmoginov, and L.C.Chen.
\newblock {MobileNetV2: Inverted Residuals and Linear Bottlenecks}.
\newblock In {\em {IEEE Conference on Computer Vision and Pattern Recognition
  (CVPR)}}, pages 4510--4520, 2018.

\bibitem{DenseNet}
G.~Huang, Z.~Liu, L.~van~der Maaten, and K.~Q. Weinberger.
\newblock {Densely Connected Convolutional Networks}.
\newblock In {\em {IEEE Conference on Computer Vision and Pattern Recognition
  (CVPR)}}, pages 2261--2269, 2017.

\bibitem{Inception}
C.~Szegedy, V.~Vanhoucke, S.~Ioffe, J.~Shlens, and W.~Zbigniew.
\newblock {Rethinking the Inception Architecture for Computer Vision}.
\newblock In {\em {IEEE Conference on Computer Vision and Pattern Recognition
  (CVPR)}}, pages 2818--2826, 2016.

\bibitem{InceptionResNet}
C.~Szegedy, S.~Ioffe, V.~Vanhoucke, and A.~A. Alemi.
\newblock {Inception-v4, Inception-ResNet and the Impact of Residual
  Connections on Learning}.
\newblock In {\em {Thirty-First AAAI Conference on Artificial Intelligence
  (AAAI)}}, pages 4278--4284, 2017.

\bibitem{Xception}
F.~Chollet.
\newblock {Xception: Deep Learning with Depthwise Separable Convolutions}.
\newblock In {\em {IEEE Conference on Computer Vision and Pattern Recognition
  (CVPR)}}, pages 1251--1258, 2017.

\bibitem{NASNet}
B.~Zoph, V.~Vasudevan, J.~Shlens, and Q.~V. Le.
\newblock {Learning Transferable Architectures for Scalable Image Recognition}.
\newblock In {\em {IEEE Conference on Computer Vision and Pattern Recognition
  (CVPR)}}, pages 8697--8710, 2018.

\bibitem{intESN}
D.~Kleyko, E.~P. Frady, M.~Kheffache, and E.~Osipov.
\newblock {Integer Echo State Networks: Efficient Reservoir Computing for
  Digital Hardware}.
\newblock {\em {IEEE Transactions on Neural Networks and Learning Systems}},
  33(4):1688--1701, 2022.

\bibitem{YilmazMachine2015}
O.~Yilmaz.
\newblock {Machine Learning Using Cellular Automata Based Feature Expansion and
  Reservoir Computing}.
\newblock {\em {Journal of Cellular Automata}}, 10(5-6):435--472, 2015.

\bibitem{YilmazSymbolic2015}
O.~Yilmaz.
\newblock {Symbolic Computation Using Cellular Automata-Based Hyperdimensional
  Computing}.
\newblock {\em {Neural Computation}}, 27(12):2661--2692, 2015.

\bibitem{mcdonald2017reservoir}
N.~McDonald.
\newblock {Reservoir Computing \& Extreme Learning Machines using Pairs of
  Cellular Automata Rules}.
\newblock In {\em International Joint Conference on Neural Networks (IJCNN)},
  pages 2429--2436, 2017.

\bibitem{KleykoBrainsCA2017}
D.~Kleyko and E.~Osipov.
\newblock {No Two Brains Are Alike: Cloning a Hyperdimensional Associative
  Memory Using Cellular Automata Computations}.
\newblock In {\em Biologically Inspired Cognitive Architectures (BICA)}, volume
  636 of {\em Advances in Intelligent Systems and Computing}, pages 91--100,
  2017.

\bibitem{SchmuckHardwareOptimizations2019}
M.~Schmuck, L.~Benini, and A.~Rahimi.
\newblock Hardware optimizations of dense binary hyperdimensional computing:
  Rematerialization of hypervectors, binarized bundling, and combinational
  associative memory.
\newblock {\em {ACM Journal on Emerging Technologies in Computing Systems}},
  15(4):1--25, 2019.

\bibitem{KleykoCA2020}
D.~Kleyko, E.~P. Frady, and F.~T. Sommer.
\newblock {Cellular Automata Can Reduce Memory Requirements of Collective-State
  Computing}.
\newblock {\em {IEEE Transactions on Neural Networks and Learning Systems}},
  33(6):2701--2713, 2022.

\bibitem{MenonEmotionHD2021}
A.~Menon, A.~Natarajan, R.~Agashe, D.~Sun, M.~Aristio, H.~Liew, Y.~S. Shao, and
  J.~M. Rabaey.
\newblock {Efficient Emotion Recognition Using Hyperdimensional Computing with
  Combinatorial Channel Encoding and Cellular Automata}.
\newblock {\em {Brain Informatics}}, 9:1--13, 2022.

\bibitem{HerscheHDMClassifier2021}
M.~Hersche, S.~Lippuner, M.~Korb, L.~Benini, and A.~Rahimi.
\newblock {Near-channel Classifier: Symbiotic Communication and Classification
  in High-dimensional Space}.
\newblock {\em {Brain Informatics}}, 8:1--15, 2021.

\bibitem{KleykoDecoding2022}
D.~Kleyko, C.~Bybee, P.-C. Huang, C.~J. Kymn, B.~A. Olshausen, F.~T. Sommer,
  and E.~P. Frady.
\newblock {Efficient Decoding of Compositional Structure in Holistic
  Representations}.
\newblock {\em {arXiv}}, 2022.

\bibitem{RachkovskijAnalogical2004}
D.~A. Rachkovskij.
\newblock {Some Approaches to Analogical Mapping with Structure Sensitive
  Distributed Representations}.
\newblock {\em Journal of Experimental and Theoretical Artificial
  Intelligence}, 16(3):125--145, 2004.

\bibitem{RachkovskijAnalogy2012}
D.~A. Rachkovskij and S.~V. Slipchenko.
\newblock {Similarity-based Retrieval with Structure-Sensitive Sparse Binary
  Distributed Representations}.
\newblock {\em Computational Intelligence}, 28(1):106--129, 2012.

\bibitem{OsipovHD_FSA2017}
E.~Osipov, D.~Kleyko, and A.~Legalov.
\newblock {Associative Synthesis of Finite State Automata Model of a Controlled
  Object with Hyperdimensional Computing}.
\newblock In {\em Annual Conference of the IEEE Industrial Electronics Society
  (IECON)}, pages 3276--3281, 2017.

\bibitem{KleykoABF2020}
D.~Kleyko, A.~Rahimi, R.~W. Gayler, and E.~Osipov.
\newblock {Autoscaling Bloom Filter: Controlling Trade-off Between True and
  False Positives}.
\newblock {\em Neural Computing and Applications}, 32:3675--3684, 2020.

\bibitem{YerxaUCBHD_FSA2018}
T.~Yerxa, A.~Anderson, and E.~Weiss.
\newblock {The Hyperdimensional Stack Machine}.
\newblock In {\em Cognitive Computing}, pages 1--2, 2018.

\bibitem{PashchenkoSubstring2020}
D.~V. Pashchenko, D.~A. Trokoz, A.~I. Martyshkin, M.~P. Sinev, and B.~L.
  Svistunov.
\newblock {Search for a Substring of Characters using the Theory of
  Non-deterministic Finite Automata and Vector-Character Architecture}.
\newblock {\em Bulletin of Electrical Engineering and Informatics},
  9(3):1238--1250, 2020.

\bibitem{KleykoPermuted2016}
D.~Kleyko, E.~Osipov, and R.~W. Gayler.
\newblock {Recognizing Permuted Words with Vector Symbolic Architectures: A
  Cambridge Test for Machines}.
\newblock {\em {Procedia Computer Science}}, 88:169--175, 2016.

\bibitem{JoshiNgrams2016}
A.~Joshi, J.~T. Halseth, and P.~Kanerva.
\newblock {Language Geometry Using Random Indexing}.
\newblock In {\em {International Symposium on Quantum Interaction (QI)}}, pages
  265--274, 2016.

\bibitem{RachkovskijRecursiveBinding2022}
D.~A. Rachkovskij and D.~Kleyko.
\newblock {Recursive Binding for Similarity-Preserving Hypervector
  Representations of Sequences}.
\newblock In {\em {International Joint Conference on Neural Networks (IJCNN)}},
  pages 1--8, 2022.

\bibitem{KleykoMACOM2012}
D.~Kleyko, N.~Lyamin, E.~Osipov, and L.~Riliskis.
\newblock {Dependable MAC Layer Architecture based on Holographic Data
  Representation using Hyper-Dimensional Binary Spatter Codes}.
\newblock In {\em Multiple Access Communications (MACOM)}, volume 7642 of {\em
  Lecture Notes in Computer Science}, pages 134--145, 2012.

\bibitem{KimHDM2018}
H.-S. Kim.
\newblock {HDM: Hyper-Dimensional Modulation for Robust Low-Power
  Communications}.
\newblock In {\em IEEE International Conference on Communications (ICC)}, pages
  1--6, 2018.

\bibitem{WirelessGuirado2022}
R.~Guirado, A.~Rahimi, G.~Karunaratne, E.~Alarcón, A.~Sebastian, and
  S.~Abadal.
\newblock {Wireless On-Chip Communications for Scalable In-memory
  Hyperdimensional Computing}.
\newblock In {\em {International Joint Conference on Neural Networks (IJCNN)}},
  pages 1--8, 2022.

\bibitem{RC09}
{M. Lukosevicius and H. Jaeger}.
\newblock {Reservoir computing approaches to recurrent neural network
  training}.
\newblock {\em Computer Science Review}, {3}({3}):{127--149}, {2009}.

\bibitem{TanakaPhysicalRC2019}
G.~Tanaka, T.~Yamane, J.~B. H{\'e}roux, R.~Nakane, N.~Kanazawa, S.~Takeda,
  H.~Numata, D.~Nakano, and A.~Hirose.
\newblock {Recent Advances in Physical Reservoir Computing: A review}.
\newblock {\em {Neural Networks}}, 115:100--123, 2019.

\end{thebibliography}

\end{document}